\documentclass{article} 
\usepackage{iclr2021_conference,times}

\usepackage{amsmath,amsfonts,bm}

\def\eqref#1{equation~\ref{#1}}

\def\1{\bm{1}}

\DeclareMathAlphabet{\mathsfit}{\encodingdefault}{\sfdefault}{m}{sl}
\SetMathAlphabet{\mathsfit}{bold}{\encodingdefault}{\sfdefault}{bx}{n}

\newcommand{\E}{\mathbb{E}}

\newcommand{\R}{\mathbb{R}}

\newcommand{\argmaxtheta}[0]{\underset{\theta}{\operatorname{argmax}}~}

\usepackage{hyperref}
\usepackage{url}
\usepackage{graphicx}
\usepackage[ruled,vlined]{algorithm2e}
\usepackage{multirow}
\usepackage{amssymb}
\usepackage{pifont}
\usepackage{booktabs}
\usepackage{tabularx}
\usepackage{xspace}
\usepackage{multirow}
\usepackage{array}
\usepackage{wrapfig}
\newcommand{\PreserveBackslash}[1]{\let\temp=\\#1\let\\=\temp}
\newcolumntype{C}[1]{>{\PreserveBackslash\centering}p{#1}}
\newcolumntype{R}[1]{>{\PreserveBackslash\raggedleft}p{#1}}
\newcolumntype{L}[1]{>{\PreserveBackslash\raggedright}p{#1}}

\definecolor{red}{RGB}{160,0,0}
\definecolor{green}{RGB}{0,150,0}
\newcommand{\cmark}{\textcolor{green}{\ding{51}}}
\newcommand{\xmark}{\textcolor{red}{\ding{55}}}

\newcommand*{\methodname}{VERA}

\definecolor{mydarkblue}{HTML}{001373}

\hypersetup{
    pdftitle={},
    pdfauthor={},
    pdfsubject={},
    pdfkeywords={},
    pdfborder=0 0 0,
    pdfpagemode=UseNone,
    colorlinks=true,
    linkcolor=mydarkblue,
    citecolor=mydarkblue,
    filecolor=mydarkblue,
    urlcolor=mydarkblue,
    pdfview=FitH
}

\title{No MCMC for me: Amortized sampling for fast and stable training of energy-based models}

\iclrfinalcopy

\newcommand*\samethanks[1][\value{footnote}]{\footnotemark[#1]}
\author{Will Grathwohl\thanks{Equal Contribution. Code available at \href{https://github.com/wgrathwohl/VERA}{\nolinkurl{github.com/wgrathwohl/VERA}}}\\
University of Toronto \& Vector Institute\\
Google Research\\
\texttt{wgrathwohl@cs.toronto.edu} \\
\And
Jacob Kelly\samethanks \\
University of Toronto \& Vector Institute\\
\texttt{jkelly@cs.toronto.edu} \\
\And
Milad Hashemi\\
Google Research\\
\texttt{miladh@google.com} \\
\And
Mohammad Norouzi \& Kevin Swersky\\
Google Research \\
\texttt{\{mnorouzi, kswersky\}@google.com} \\
\And
David Duvenaud \\
University of Toronto \& Vector Institute \\
\texttt{duvenaud@cs.toronto.edu} \\
}

\begin{document}

\maketitle

\begin{abstract}
Energy-Based Models (EBMs) present a flexible and appealing way to represent uncertainty. Despite recent advances, training EBMs on high-dimensional data remains a challenging problem as the state-of-the-art approaches are costly, unstable, and require considerable tuning and domain expertise to apply successfully. In this work we present a simple method for training EBMs at scale which uses an entropy-regularized generator to amortize the MCMC sampling typically used in EBM training. We improve upon prior MCMC-based entropy regularization methods with a fast variational approximation.
We demonstrate the effectiveness of our approach by using it to train tractable likelihood models.
Next, we apply our estimator to the recently proposed Joint Energy Model (JEM), where we match the original performance with faster and stable training.
This allows us to extend JEM models to semi-supervised classification on tabular data from a variety of continuous domains.


\end{abstract}

\section{Introduction}

Energy-Based Models (EBMs) have recently regained popularity within machine learning, partly inspired by the impressive results of~\citet{du2019implicit} and \citet{song2020improved}
on large-scale image generation. Beyond image generation, EBMs have also been successfully applied to a wide variety of applications including: out-of-distribution detection~\citep{grathwohl2019your, du2019implicit, song2018learning}, adversarial robustness~\citep{grathwohl2019your, hill2020stochastic, du2019implicit}, reliable classification~\citep{grathwohl2019your, liu2020hybrid} and semi-supervised learning~\citep{song2018learning, zhaojoint}. Strikingly, these EBM approaches outperform alternative classes of generative models and rival hand-tailored solutions on each task.

\begin{table}[t]
\centering
\small
\renewcommand{\arraystretch}{1.3}
\begin{tabular}{ l  C{0.4cm} @{\hspace{.2cm}} C{1.3cm} @{\hspace{.3cm}} C{1.5cm} @{\hspace{.3cm}} C{1.1cm} @{\hspace{.3cm}} C{1.5cm} @{}@{\hspace{.3cm}} C{1.4cm} @{}} 
Training Method & Fast & Stable training & High dimensions & No aux. model & Unrestricted architecture & Approximates likelihood \\
\midrule
Markov chain Monte Carlo & \xmark & \xmark & \cmark & \cmark &  \cmark & \cmark \\
Score Matching Approaches & \cmark & \xmark & \cmark &  \cmark & \xmark & \xmark\\
Noise Contrastive Approaches & \cmark & \cmark & \xmark &  \xmark & \cmark & \xmark \\
\midrule
\methodname{} (ours) & \cmark & \cmark & \cmark & \xmark & \cmark & \cmark \\
\end{tabular}
\vspace{-0.1cm}
\caption{Features of EBM training approaches. Trade-offs must be made when training unnormalized models and no approach to date satisfies all of these properties.}
\label{tab:comparison}
\end{table}

Despite progress, training EBMs is still a challenging task. As shown in Table~\ref{tab:comparison}, existing training methods are all deficient in at least one important practical aspect.
Markov chain Monte Carlo (MCMC) methods
are slow and unstable during training 
~\citep{nijkamp2019anatomy, grathwohllearning}.
Score matching mechanisms, which minimize alternative divergences are also unstable and most methods cannot work with discontinuous nonlinearities (such as ReLU)~\citep{song2019generative,hyvarinen2005estimation,song2020sliced, pang2020efficient, grathwohllearning, vincent2011connection}.
Noise contrastive approaches, which learn energy functions through density ratio estimation, typically don't scale well to high-dimensional data~\citep{gao2020flow, rhodes2020telescoping, gutmann2010noise, ceylan2018conditional}.

In this work, we present a simple method for training EBMs which performs as well as previous methods while being faster and substantially easier to tune.
Our method is based on reinterpreting maximum likelihood as a bi-level variational optimization problem, which has been explored in the past for EBM training~\citep{dai2019exponential}.
This perspective allows us to amortize away MCMC sampling into a GAN-style generator which is encouraged to have high entropy. We accomplish this with a novel approach to entropy regularization based on a fast variational approximation. This leads to the method we call Variational Entropy Regularized Approximate maximum likelihood (VERA).



\begin{figure}[t]
\centering
\includegraphics[height=.3\textwidth]{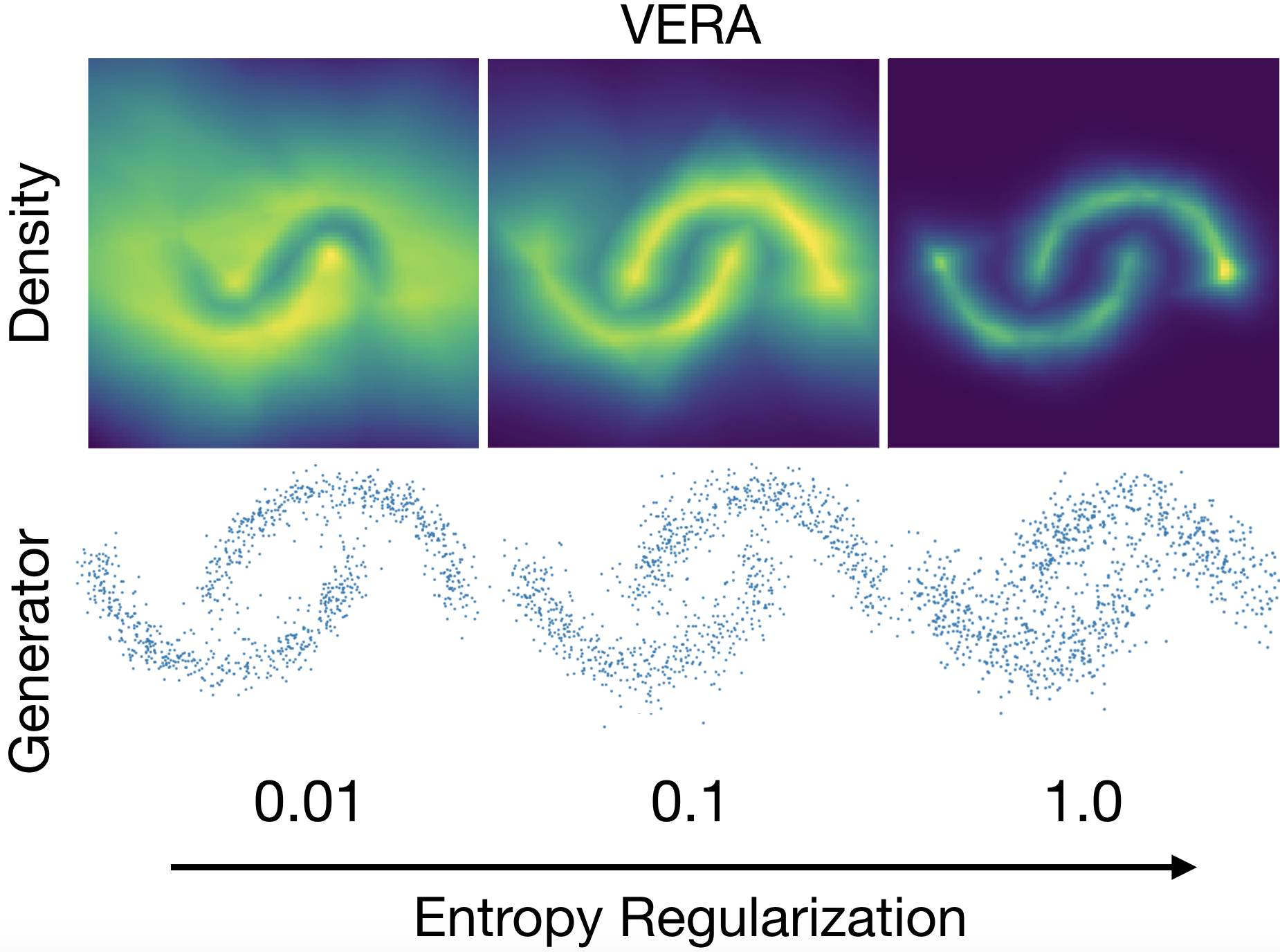}
\medskip
\includegraphics[height=.3\textwidth]{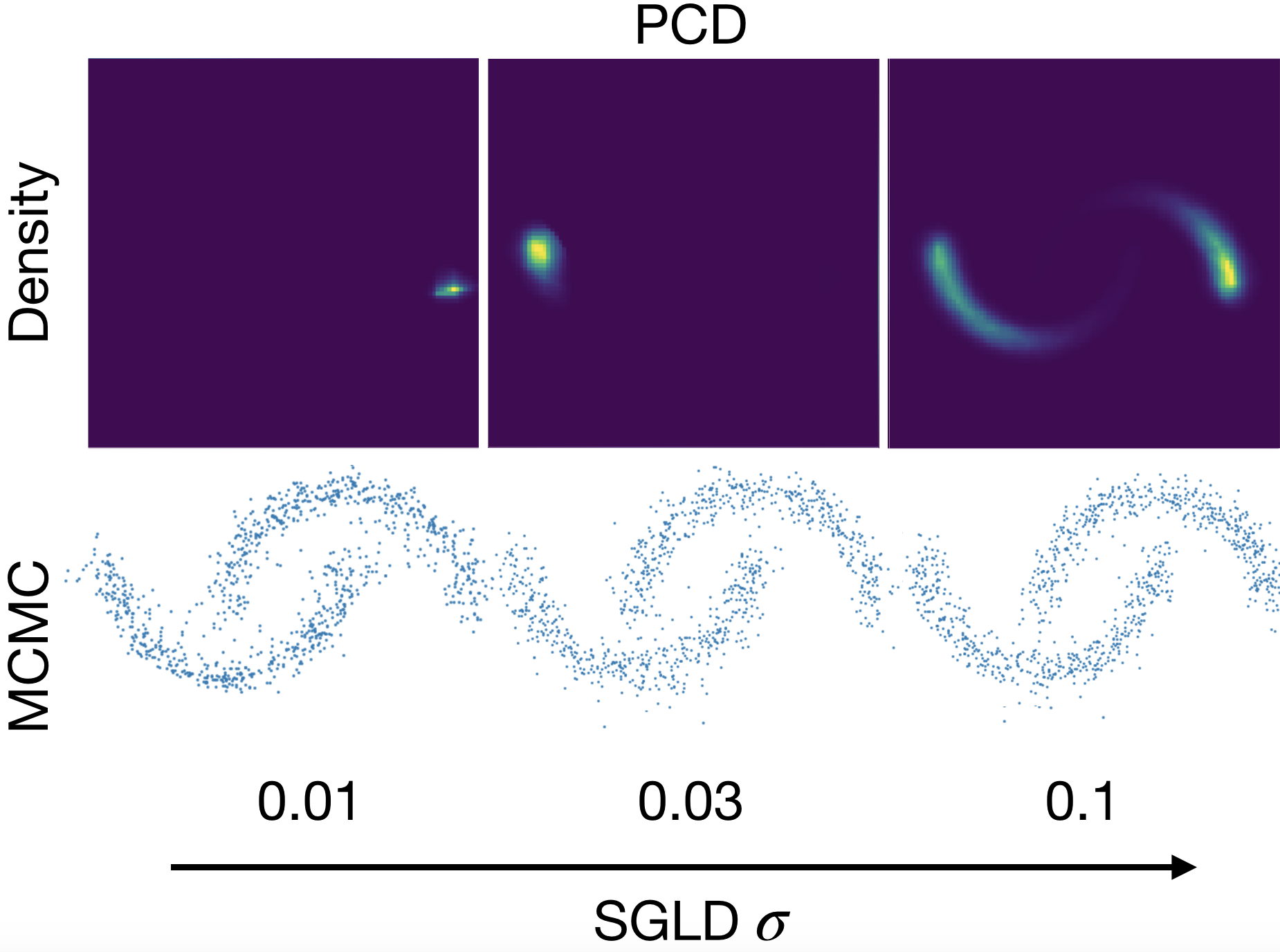}
\medskip
\includegraphics[height=.3\textwidth]{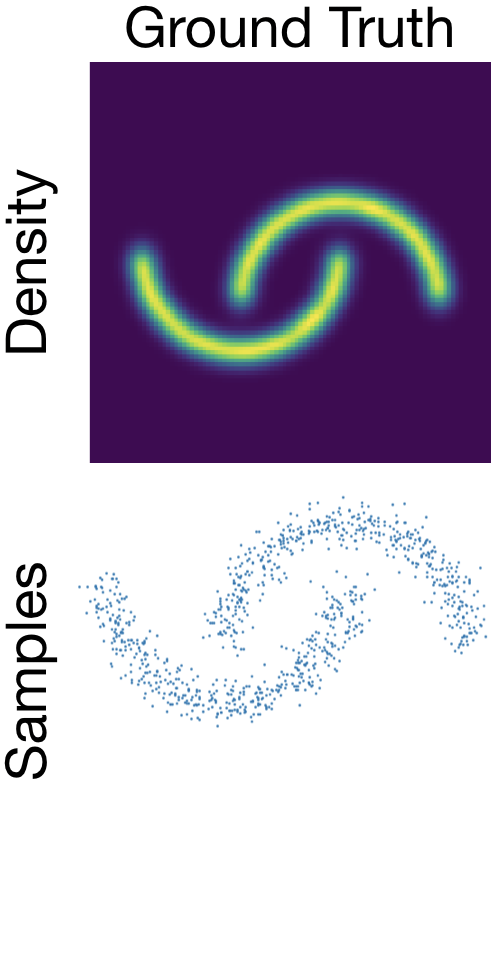}
    \vspace{-2em}
    \caption{Comparison of EBMs trained with \methodname{} and PCD. We see that as entropy regularization goes to $1$, the density becomes more accurate. For PCD, all samplers produce high quality samples, but low-quality density models as the distribution of MCMC samples may be arbitrarily far away from the model density.}
    \label{fig:toy}
    \vspace{-1.7em}
\end{figure}



Concretely, we make the following contributions:
\begin{itemize}
\item We improve the MCMC-based entropy regularizer of \citet{dieng2019prescribed} with a parallelizable variational approximation.
\item We show that an entropy-regularized generator can be used to produce a variational bound on the EBM likelihood which can be optimized more easily than MCMC-based estimators.
\item We demonstrate that models trained in this way achieve much higher likelihoods than methods trained with alternative EBM training procedures.
\item We show that our approach stabilizes and accelerates the training of recently proposed Joint Energy Models~\citep{grathwohl2019your}.

\item We show that the stabilization of our approach allows us to use JEM for semi-supervised learning, outperforming virtual adversarial training when little prior domain knowledge is available (e.g., for tabular data).

\end{itemize}

\section{Energy Based Models}
An energy-based model (EBM) is any model which parameterizes a density as
\begin{align}
    p_\theta(x) = \frac{e^{f_\theta(x)}}{Z(\theta)} \label{eq:ebm}
\end{align}
where $f_\theta: \mathbb{R}^D \rightarrow \mathbb{R}$ and $Z(\theta) = \int e^{f_\theta(x)}\mathrm{d} x$ is the normalizing constant which is not explicitly modeled.
Any probability distribution can be represented in this way for some $f_\theta$.
The energy-based parameterization has been used widely for its flexibility, ease of incorporating known structure, and relationship to physical systems common in chemistry, biology, and physics~\citep{ingraham2019learning, du2020energy, noe2019boltzmann}.

The above properties make EBMs an appealing model class, but because they are unnormalized many tasks which are simple for alternative model classes become challenging for EBMs. For example, exact samples cannot be drawn and likelihoods cannot be exactly computed (or even lower-bounded). This makes training EBMs challenging as we cannot simply train them to maximize likelihood.
The most popular approach to train EBMs is to approximate the gradient of the maximum likelihood objective. This gradient can be written as:
\begin{align}
    \nabla_\theta \log p_\theta(x) = \nabla_\theta f_\theta(x) - \E_{p_\theta(x')}[\nabla_\theta f_\theta(x')].
    \label{eq:approxML}
\end{align}
MCMC techniques are used to approximately generate samples from $p_\theta(x)$~\citep{tieleman2008training}. Practically, this approach suffers from poor stability and computational challenges from sequential sampling. Many tricks have been developed to overcome these issues~\citep{du2019implicit}, but they largely still persist. Alternative estimators have been proposed to circumvent these challenges, including score matching~\citep{hyvarinen2005estimation}, noise contrastive estimation~\citep{gutmann2010noise}, and variants thereof. These suffer from their own challenges in scaling to high dimensional data, and sacrifice the statistical efficiency of maximum likelihood. 

In Figure~\ref{fig:toy} we visualize densities learned with our approach and Persistent Contrastive Divergence~\citep{tieleman2008training} (PCD) training. As we see, the sample quality of the PCD models is quite high but the learned density models do not match the true model. This is due to accrued bias in the gradient estimator from approximate MCMC sampling~\citep{grathwohllearning}. Prior work~\citep{nijkamp2019learning} has argued that this objective actually encourages the approximate MCMC samples to match the data rather than the density model. Conversely, we see that our approach (with proper entropy regularization) recovers a high quality model.






\section{Variational Maximum Likelihood}
\label{sec:dual}
We seek the energy function which maximizes likelihood given in Equation \ref{eq:ebm}. 
We examine the intractable component of the log-likelihood, the log partition-function $\log Z(\theta) = \log \int e^{f_\theta(x)}\mathrm{d}x$. We can re-write this quantity as the optimum of
\begin{align}
    \log Z(\theta) = \max_{q} \E_{q(x)}[f_\theta(x)] + H(q) \label{eq:fencheldual}
\end{align}
where $q$ is a distribution and $H(q) = -\E_{q(x)}[\log q(x)]$ denotes its entropy~\footnote{For continuous spaces, this would be the differential entropy, but we simply use entropy here for brevity.} (see the Appendix~\ref{app:dual_der} for the derivation). Plugging this into our original maximum likelihood statement we obtain:
\begin{align}
    \hat{\theta} = \argmaxtheta \left [ \E_{p_\text{data}(x)} [f_\theta(x)]
    - \max_q\Big[ \E_{q(x)} [f_\theta(x) ]
    + H(q)\Big]\right ]
    \label{eq:dual_likelihood}
\end{align}
which gives us an alternative method for training EBMs. We introduce an auxiliary sampler $q_\phi$ which we train online to optimize the inner-loop of Equation \ref{eq:dual_likelihood}. This objective was used for EBM training in \citet{kumar2019maximum, abbasnejad2019generative, dai2017calibrating},  \citet{dai2019exponential} (motivated by Fenchel Duality~\citep{wainwright2008graphical}). \citet{abbasnejad2019generative} use an implicit generative model and \citet{dai2019exponential} propose to use a sampler which is inspired by MCMC sampling from $p_\theta(x)$ and whose entropy can be computed exactly.

Below we describe our approach which utilizes the same objective with a simpler sampler and a new approach to encourage high entropy. We note that when training $p_\theta(x)$ and $q(x)$ online together, the inner maximization will not be fully optimized. This leads our training objective for $p_\theta(x)$ to be an \emph{upper} bound on $\log p_\theta(x)$. In Section \ref{sec:likelihood} we explore the impact of this fact and find that the bound is tight enough to train models that achieve high likelihood on high-dimensional data.   

\section{Method}

We now present a method for training an EBM $p_\theta(x) = e^{f_\theta(x)}/Z(\theta)$ to optimize Equation \ref{eq:dual_likelihood}. We introduce a generator distribution of the form $q_\phi(x) = \int_z q_\phi(x|z)q(z)\mathrm{d}z$ such that:
\begin{align}
    q_\phi(x|z) = \mathcal{N}(g_\psi(z), \sigma^2 I), \qquad q(z) = \mathcal{N}(0, I)
\end{align}
where $g_\psi$ is a neural network with parameters $\psi$ and thus, $\phi = \{\psi, \sigma^2\}$.
This is similar to the decoder of a variational autoencoder~\citep{kingma2013auto}. With this architecture it is easy to optimize the first and second terms of Equation \ref{eq:dual_likelihood} with reparameterization, but the entropy term requires more care.

\subsection{Entropy Regularization}
Estimating entropy or its gradients is a challenging task. Multiple, distinct approaches have been proposed in recent years based on Mutual Information estimation~\citep{kumar2019maximum}, variational upper bounds~\citep{ranganath2016hierarchical}, Denoising Autoencoders~\citep{lim2020ar}, and nearest neighbors~\citep{singh2016analysis}.

The above methods require the training of additional auxiliary models or do not scale well to high dimensions. Most relevant to this work are \citet{dieng2019prescribed, titsias2019unbiased} which present a method for encouraging generators such as ours to have high entropy by estimating $\nabla_\phi H(q_\phi)$.
The estimator takes the following form: 
\begin{align}
    \nabla_\phi H(q_\phi) &= \nabla_\phi \E_{q_\phi(x)}[\log q_\phi(x)]\nonumber \\
    &= \nabla_\phi \E_{p(z)p(\epsilon)}[\log q_\phi (x(z, \epsilon))] \qquad \text{(Reparameterize sampling)}\nonumber\\
    &=  \E_{p(z)p(\epsilon)}[\nabla_\phi \log q_\phi (x(z, \epsilon))]\nonumber\\
    &=  \E_{p(z)p(\epsilon)}[\nabla_x \log q_\phi (x(z, \epsilon))^T \nabla_\phi x(z, \epsilon)] \qquad \text{(Chain rule)}
    \label{eq:ent_est}
\end{align}
where we have written $x(z, \epsilon) = g_\psi(z) + \sigma \epsilon$.
All quantities in Equation \ref{eq:ent_est} can be easily computed except for the score-function $\nabla_x \log q_\phi(x)$. The following estimator for this quantity can be easily derived (see Appendix~\ref{app:score_fn}):
\begin{align}
    \nabla_x \log q_\phi (x) = \E_{q_\phi(z|x)}[\nabla_x \log q_\phi(x|z)] \label{eq:grad_logqx}
\end{align}
which requires samples from the posterior $q_\phi(z|x)$ to estimate. \citet{dieng2019prescribed, titsias2019unbiased} generate these samples using Hamiltonian Monte Carlo (HMC)~\citep{neal2011mcmc}, a gradient-based MCMC algorithm. As used in \citet{dieng2019prescribed}, 28 \emph{sequential} gradient computations must be made per training iteration. Since a key motivation of this work is to circumvent the costly sequential computation of MCMC sampling, this is not a favourable solution. In our work we propose a more efficient solution that we find works just as well empirically. 

\subsection{Variational Approximation with Importance Sampling}
We propose to replace HMC sampling of $q_\phi(z|x)$ with a variational approximation ${\xi(z\mid z_0) \approx q_\phi(z\mid x)}$ where $z_0$ is a conditioning variable we will define shortly. We can use this approximation with self-normalized importance sampling to estimate
\begin{eqnarray}
    \nabla_x \log q_\phi (x)
    &=& \E_{q_\phi(z\mid x)} [\nabla_x \log q_\phi(x \mid z)]  \nonumber  \\
    &=& \E_{p(z_0)q_\phi(z\mid x)} [\nabla_x \log q_\phi(x \mid z)]  \nonumber  \\
    &=& \E_{p(z_0)\xi(z \mid z_0)} \left [ \frac{q_\phi(z\mid x)}{\xi(z \mid z_0)} \nabla_x \log q_\phi(x \mid z) \right]\nonumber \\
    &=& \E_{p(z_0)\xi(z \mid z_0)} \left [ \frac{q_\phi(z, x)}{q_\phi(x)\xi(z \mid z_0)} \nabla_x \log q_\phi(x \mid z) \right]\nonumber \\
    &\approx& \sum_{i=1}^{k} \frac{w_i}{\sum_{j=1}^{k} w_j} \nabla_x \log q_\phi(x \mid z_i) \equiv \widetilde{\nabla}_x\log q_\phi(x; \{z_i\}_{i=1}^k, z_0)\label{eq:variational_snis}
\end{eqnarray}
where $\{z_i\}_{i=1}^k \sim \xi(z\mid z_0)$ and $w_i \equiv \frac{q_\phi(z_i, x)}{\xi(z_i \mid z_0)}$. We use $k=20$ importance samples for all experiments in this work. This approximation holds for any conditioning information we would like to use. To choose this, let us consider how samples $x\sim q_\phi(x)$ are drawn. We first sample $z_0 \sim \mathcal{N}(0, I)$ and then $x\sim q_\phi(x\mid z_0)$. In our estimator we want a variational approximation to $q_\phi(z\mid x)$ and by construction, $z_0$ is a sample from this distribution. For this reason we let our variational approximation be 
\begin{align}
\xi_\eta(z\mid z_0) = \mathcal{N}(z \mid z_0, \eta^2 I),
\end{align}
or simply a diagonal Gaussian centered at the $z_0$ which generated $x$. For this approximation to be useful we must tune the variance $\eta^2$. We do this by optimizing the standard Evidence Lower-Bound at every training iteration
\begin{align}
    \mathcal{L}_{\mathrm{ELBO}}(\eta ; z_0, x) &= \mathbb{E}_{\xi_\eta(z \mid z_0)}\left [ \log(q_\phi(x \mid z)) + \log(q_\phi(z))\right ] + H(\xi_\eta(z \mid z_0)). \label{eq:iselbo}
\end{align}

We then use $\xi_\eta(z\mid z_0)$ to approximate $\nabla_x \log q_\phi(x)$ which we plug into Equation \ref{eq:ent_est} to estimate $\nabla_\phi H(q_\phi)$ for training our generator. A full derivation and discussion can be found in Appendix~\ref{app:ent_est}.

Combining the tools presented above we arrive at our proposed method which we call Variational Entropy Regularized Approximate maximum likelihood (\methodname{}), outlined in Algorithm \ref{alg:vera}. We found it helpful to further add an $\ell_2$-regularizer with weight $0.1$ to the gradient of our model's likelihood as in \citet{kumar2019maximum}. In some of our larger-scale experiments we reduced the weight of the entropy regularizer as in \citet{dieng2019prescribed}. We refer to the entropy regularizer weight as $\lambda$.

\begin{algorithm}[h]
\SetAlgoLined
\SetNoFillComment
\DontPrintSemicolon
\SetKwInOut{Input}{Input}
\SetKwInOut{Output}{Output}
\Input{EBM $p_\theta(x)\propto e^{f_\theta(x)}$, \text{ generator } $q_\phi(x, z)$, \text{ approximate posterior } $\xi_\eta(z|z_0)$, \text{ entropy weight } $\lambda$, \text{ gradient penalty } $\gamma=.1$}
\Output{Parameters $\theta$ such that $p_\theta \approx p$}
\caption{\methodname{} Training}\label{alg:vera}
\While{True}{
    Sample mini-batch $x$, and generate mini-batch $x_g, z_0 \sim q_\phi(x, z)$\\
    Compute $\mathcal{L}_{\text{ELBO}}(\eta; z_0, x_g)$ and update $\eta$ \tcp*{Update posterior}
    Compute $\log f_\theta(x) - \log f_\theta(x_g) + \gamma ||\nabla_x \log f_\theta(x)||^2$ and update $\theta$ \tcp*{Update EBM}
    Sample $\{z_i\}_{i=1}^k \sim \xi_\eta(z|z_0)$\\
    Compute $s = \widetilde{\nabla}_x\log q_\phi(x; \{z_i\}_{i=1}^k, z_0)$ \tcp*{Estimate score fn (Eq.\ref{eq:variational_snis})}
    Compute $g = s^T \nabla_\phi x_g$ \tcp*{Estimate entropy gradient (Eq.\ref{eq:ent_est})}
    Update $\phi$ with $\nabla_\phi \log f_\theta(x_g) + \lambda g$ \tcp*[f]{Update generator}
}
\end{algorithm}


\section{EBM Training Experiments}




We present results training various models with \methodname{} and related approaches. In Figure~\ref{fig:toy} we visualize the impact of our generator's entropy on the learned density model and compare this with MCMC sampling used in PCD learning. In Section~\ref{sec:likelihood}, we explore this quantitatively by training tractable models and evaluating with test-likelihood. In Section~\ref{sec:bias} we explore the bias of our entropy gradient estimator and the estimator's effect on capturing modes.

\subsection{Fitting Tractable Models}
\label{sec:likelihood}



\begin{wrapfigure}{R}{0.5\textwidth}
  \begin{center}
\newcommand*{\mypng}[1]{\includegraphics[width=2.7cm]{fig2/#1}}%
\setlength\tabcolsep{2pt}
\begin{tabular}{m{1.1cm}m{2.7cm}m{2.7cm}}
Method & \phantom{mi}Exact Samples & Generator Samples \\
\midrule
\vspace{-3em}VERA ($\lambda = 1$) & \mypng{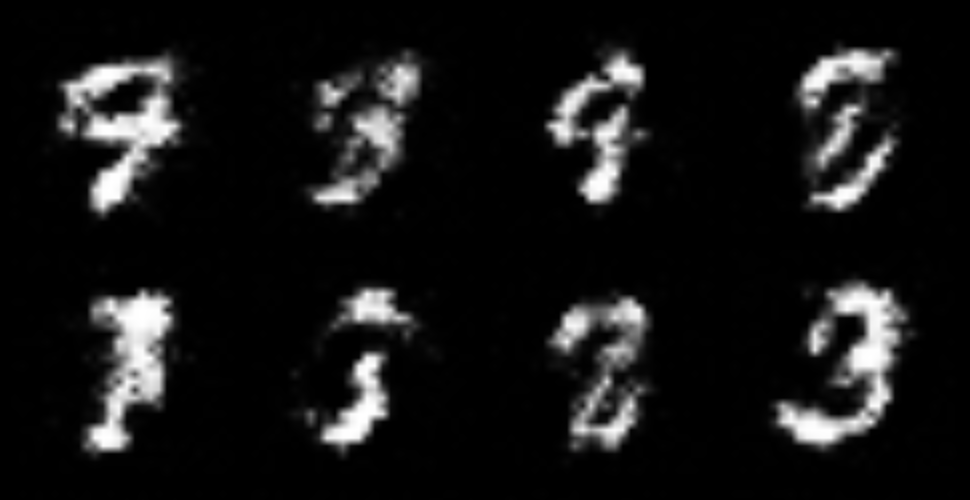} & \mypng{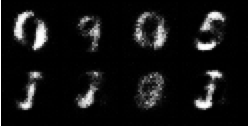} \\
\vspace{-3em}VERA ($\lambda = 0$) & \mypng{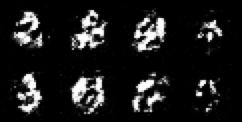} & \mypng{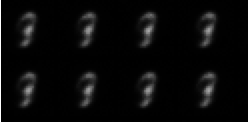} \\
MEG & \mypng{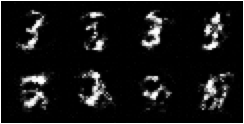} & \mypng{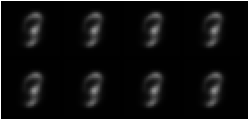} \\
PCD & \mypng{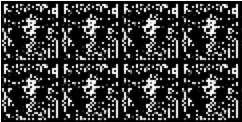} & \mypng{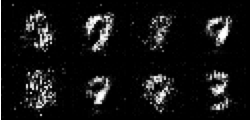} \\
\hspace{-.75em} CoopNet & \mypng{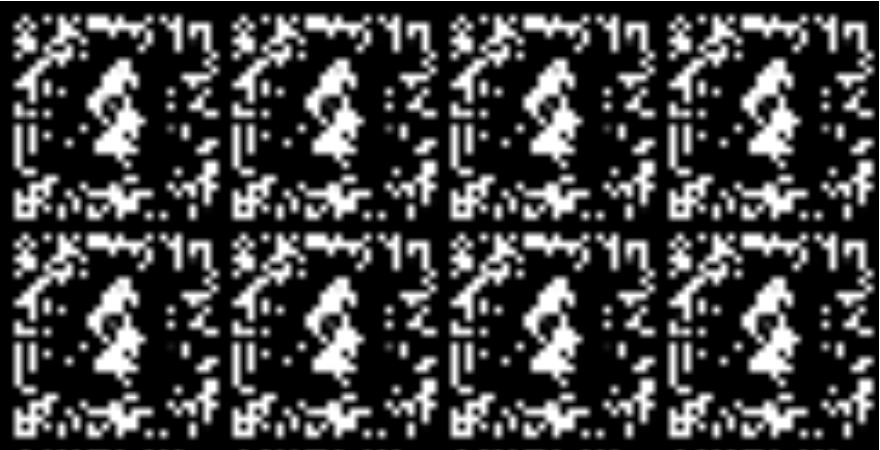} & \mypng{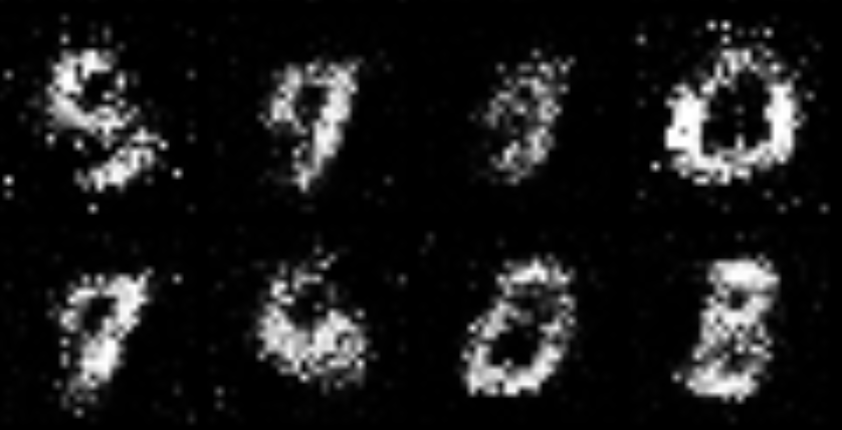}  \\
SSM & \mypng{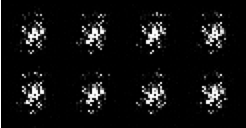} & \phantom{m}Not Applicable
\end{tabular}
  \end{center}
  \caption{\emph{Left:} Exact samples from NICE model trained with various methods. \emph{Right:} Approximate samples used for training. For \methodname{}, MEG, and CoopNet, these come from the generator, for PCD these are approximate MCMC samples.}
  \label{fig:nice_exact}
  \vspace{-2em}
\end{wrapfigure}

Optimizing the generator in \methodname{} training minimizes a variational \emph{upper} bound on the likelihood of data under our model. If this bound is not sufficiently tight, then training the model to maximize this bound will not actually improve likelihood. To demonstrate the \methodname{} bound is tight enough to train large-scale models we train NICE models~\citep{dinh2014nice} on the MNIST dataset. NICE is a normalizing flow~\citep{rezende2015variational} model -- a flexible density estimator which enables both exact likelihood computation and exact sampling. We can train this model with \methodname{} (which does not require either of these abilities), evaluate the learned model using likelihood, and generate exact samples from the trained models. Full experimental details\footnote{This experiment follows the NICE experiment in \citet{song2020sliced} and was based on their \href{https://github.com/ermongroup/sliced_score_matching}{implementation}.} can be found in Appendix~\ref{app:nice} and additional results can be found in Appendix~\ref{app:mog}.

We compare the performance of \methodname{} with maximum likelihood training as well as a number of approaches for training unnormalized models; Maximum Entropy Generators (MEG)~\citep{kumar2019maximum}, Persistent Contrastive Divergence (PCD), Sliced Score Matching  (SSM)~\citep{song2020sliced}, Denoising Score Matching  (DSM)~\citep{vincent2011connection}, Curvature Propagation (CP-SM)~\citep{martens2012estimating}, and CoopNets~\citep{xie2018cooperative}. As an ablation we also train \methodname{} with the HMC-based entropy regularizer of \citet{dieng2019prescribed}, denoted \methodname{}-(HMC). Table~\ref{tab:ml_mnist} shows that \methodname{} outperforms all approaches that do not require a normalized model. 

Figure~\ref{fig:nice_exact} shows exact samples from our NICE models. For PCD we can see (as observed in Figure~\ref{fig:toy}) that while the approximate MCMC samples resemble the data distribution, the true samples from the model do not. This is further reflected in the reported likelihood value which falls behind all methods besides DSM. CoopNets perform better than PCD, but exhibit the same behavior of generator samples resembling the data, but not matching true samples. We attribute this behavior to the method's reliance on MCMC sampling.

Conversely, models trained with \methodname{} generate coherent and diverse samples which reasonably capture the data distribution. We also see that the samples from the learned generator much more closely match true samples from the NICE model than PCD and MEG. When we remove the entropy regularizer from the generator ($\lambda = 0.0$) we observe a considerable decrease in likelihood and we find that the generator samples are far less diverse and do not match exact samples at all. Intriguingly, entropy-free \methodname{} outperforms most other methods. We believe this is because even without the entropy regularizer we are still optimizing a (weak) bound on likelihood. Conversely, the score-matching methods minimize an alternative divergence which will not necessarily correlate well with likelihood. Further, Figure~\ref{fig:nice_exact} shows that MEG performs on par with entropy-free \methodname{} indicating that the Mutual Information-based entropy estimator may not be accurate enough in high dimensions to encourage high entropy generators.


\begin{table}[h]
    \centering
    \begin{tabular}{@{}l|llccccccc@{}}
    \toprule
    Maximum & \multicolumn{2}{c}{\methodname{}} &  \methodname{} (HMC) & \multirow{2}{*}{MEG} &  \multirow{2}{*}{PCD} &  \multirow{2}{*}{SSM} &  \multirow{2}{*}{DSM} &  \multirow{2}{*}{CP-SM} &  \multirow{2}{*}{CoopNet} \\
     Likelihood               & $\lambda = 1.0$ & $\lambda = 0.0$  & $\lambda = 1.0$     &         &      &        &  \\           
    \midrule
    -791 &  \textbf{-1138} & -1214 & -1165 & -1219  & -4207 & -2039 & -4363 & -1517 & -1465\\
    \bottomrule
    \end{tabular}
    \vspace{-.6em}
    \caption{Fitting NICE models using various learning approaches for unnormalized models. 
    Results for SSM, DCM, CP-SM taken from \citet{song2020sliced}. 
    }
    \vspace{-1.2em}
    \label{tab:ml_mnist}
\end{table}



\subsection{Understanding Our Entropy Regularizer}
\label{sec:bias}

\begin{wrapfigure}{R}{0.4\textwidth}
  \begin{center}
  \vspace{-1.7em}
    \includegraphics[width=.35\textwidth, clip, trim=0cm .5cm .8cm 1.1cm]{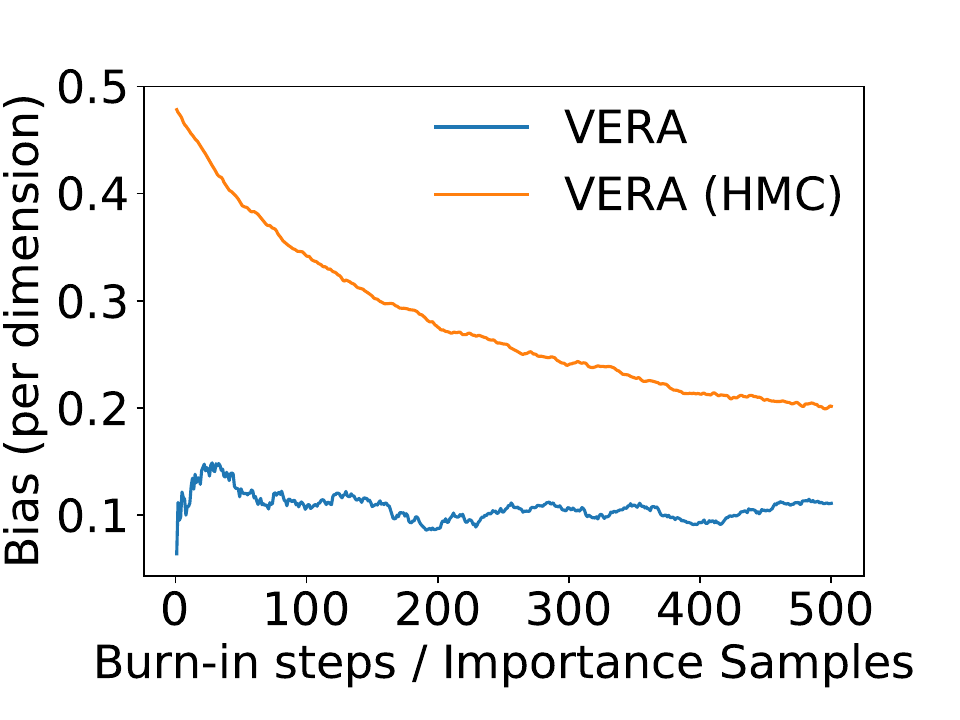}\\
    \includegraphics[width=.35\textwidth, clip, trim=0cm .5cm .8cm 1.1cm]{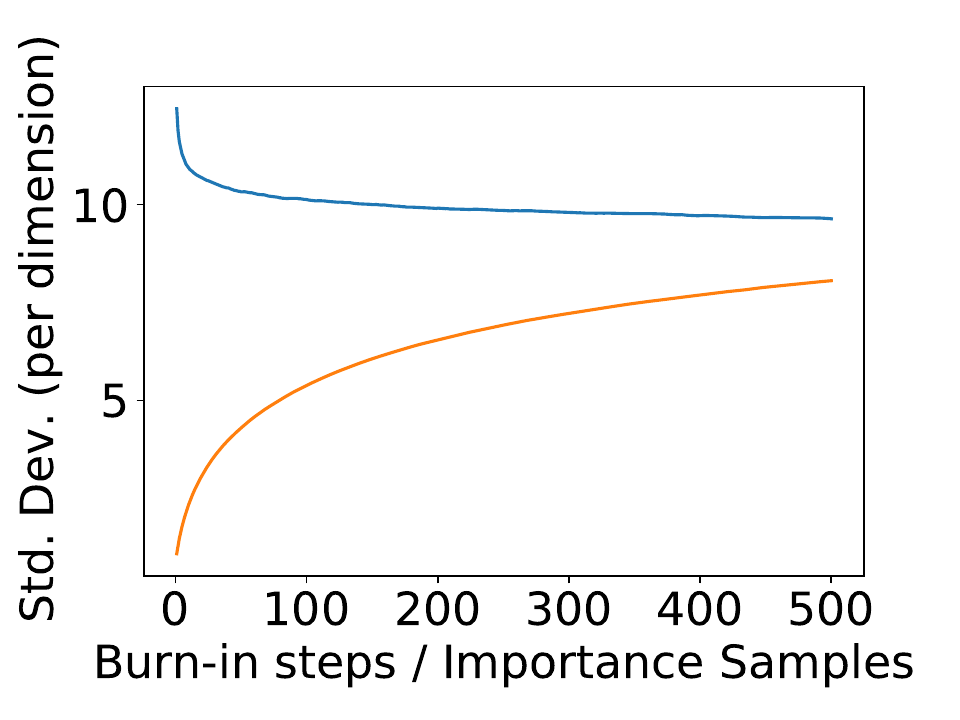}
  \end{center}
  \vspace{-1em}
  \caption{Bias (top) and standard deviation (bottom), both per dimension, of the score function estimator using HMC and our proposed importance sampling scheme.}
  \vspace{-1em}
  \label{fig:bias}
\end{wrapfigure}

In Figure~\ref{fig:bias}, we explore the quality of our score function estimator on a PCA \citep{tipping1999probabilistic} model fit to MNIST, a setting where we can compute the score function exactly (see Appendix~\ref{app:bias} for details). 
The importance sampling estimator (with 20 importance samples) has somewhat larger variance than the HMC-based estimator but has a notably lower bias of $.12$. The HMC estimator using 2 burn-in steps (recommended in \citet{dieng2019prescribed}) has a bias of $.48$. Increasing the burn-in steps to 500 reduces the bias to $.20$ while increasing variance.
We find the additional variance of our estimator is remedied by mini-batch averaging and the reduced bias helps explain the improved performance in Table \ref{tab:ml_mnist}. 

Further, we compute the effective sample size~\citep{kong1992note} (ESS) of our importance sampling proposal on our CIFAR10 and MNIST models and achieve an ESS of 1.32 and 1.29, respectively using 20 importance samples. When an uninformed proposal ($\mathcal{N}(0, I)$) is used, the ESS is 1.0 for both models. This indicates our gradient estimates are informative for training. More details can be found in Appendix \ref{app:ess}.

Next, we count the number of modes captured on a dataset with 1,000 modes consisting of 3 MNIST digits stacked on top of one another~\citep{dieng2019prescribed, kumar2019maximum}. We find that both \methodname{} and \methodname{} (HMC) recover 999 modes, but training with \emph{no entropy regularization} recovers 998 modes. 
We conclude that entropy regularization is unnecessary for preventing mode collapse in this setting.

\vspace{-.1in}
\section{Applications to Joint Energy Models}
\vspace{-.1in}

Joint Energy Models (JEM)~\citep{grathwohl2019your} are an exciting application of EBMs. They reinterpret standard classifiers as EBMs and train them as such to create powerful hybrid generative/discriminative models which improve upon purely-discriminative models at out-of-distribution detection, calibration, and adversarial robustness.

Traditionally, classification tasks are solved with a function $f_\theta: \R^D \rightarrow \R^K$ which maps from the data to $K$ unconstrained real-valued outputs (where $k$ is the number of classes). This function parameterizes a distribution over labels $y$ given data $x$: ${p_\theta(y|x) = {e^{f(x)[y]}}/{\sum_{y'} e^{f_\theta(x)[y']}}}$. The same function $f_\theta$ can be used to define an EBM for the joint distribution over $x$ and $y$ as: $p_\theta(x, y) = e^{f_\theta(x)[y]}/Z(\theta)$. The label $y$ can be marginalized out to give an unconditional model $p_\theta(x) = \sum_y e^{f_\theta(x)[y]}/{Z(\theta)}$. JEM models are trained to maximize the factorized likelihood:
\begin{align}
    \log p_\theta(x, y) = \alpha\log p_\theta(y|x) + \log p_\theta(x) 
\end{align}
where $\alpha$ is a scalar which weights the two terms. The first term is optimized with cross-entropy and the second term is optimized using EBM training methods. In \citet{grathwohl2019your} PCD was used to train the second term. We train JEM models on CIFAR10, CIFAR100, and SVHN using \methodname{} instead of PCD. We examine how this change impacts accuracy, generation, training speed, and stability. Full experimental details can be found in Appendix~\ref{app:jem}.



\paragraph{Speed and Stability}
While the results presented in \citet{grathwohl2019your} are promising, training models as presented in this work is challenging. MCMC sampling can be slow and training can easily diverge. Our CIFAR10 models train 2.8x faster than the official JEM implementation\footnote{\url{https://github.com/wgrathwohl/JEM}} with the default hyper-parameters. With these default parameters JEM models would regularly diverge. To train for the reported 200 epochs training needed to be restarted multiple times and the number of MCMC steps needed to be quadrupled, greatly increasing run-time. 

Conversely, we find that \methodname{} was much more stable and our models never diverged. This allowed us to remove the additive Gaussian noise added to the data which is very important to stabilize MCMC training~\citep{grathwohl2019your, nijkamp2019anatomy, du2019implicit}.

\paragraph{Hybrid Modeling} In Tables \ref{tab:jem_clf} and \ref{tab:jem_fid} we compare the discriminative and generative performance of JEM models trained with \methodname{}, PCD (JEM), and HDGE~\citep{liu2020hybrid}. With ${\alpha=1}$ we find that \methodname{} leads to models with poor classification performance but strong generation performance. With $\alpha=100$ \methodname{} obtains stronger classification performance than the original JEM model while still having improved image generation over JEM and HDGE (evaluated with FID~\citep{heusel2017gans}). 


\begin{table}[!htb]
\vspace{-1.8em}
\begin{minipage}{.5\linewidth}
    \centering
    \medskip
\begin{tabular}{@{}l|lll@{}}
\toprule
    Model & CIFAR10 & CIFAR100 & SVHN \\
    \midrule
    Classifier & 95.8 & 74.2 & 97.7 \\
    \midrule
        JEM  & 92.9 & 72.2 & 96.6 \\
        HDGE & 96.7 & 74.5 & N/A  \\
    \midrule
        \methodname{} $\alpha = 100$ & 93.2 & 72.2 & 96.8 \\
        \methodname{} $\alpha = 1$ & 76.1 & 48.7 & 94.2\\
    \bottomrule
\end{tabular}
\vspace{-1em}
\caption{Classification on image datasets.}
\label{tab:jem_clf}

\end{minipage}\hfill
\begin{minipage}{.5\linewidth}
    \centering
    \medskip

 \begin{tabular}{@{}l|l@{}}
    \toprule
    Model & FID $\downarrow$\\
    \midrule
     JEM & 38.4\\
     HDGE & 37.6\\
     SNGAN~\citep{miyato2018spectral} & 25.50 \\
     NCSN~\citep{song2019generative} & 23.52\\
     \midrule
     \methodname{} $\alpha = 100$  & 30.5\\
     \methodname{} $\alpha = 1$  & 27.5 \\
    \bottomrule
    \end{tabular}
    
    \vspace{-.7em}
    \caption{FID on CIFAR10.}
    \label{tab:jem_fid}
\end{minipage}
\vspace{-1em}
\end{table}

\begin{figure}[h]  
\centering
\includegraphics[width=.44\textwidth]{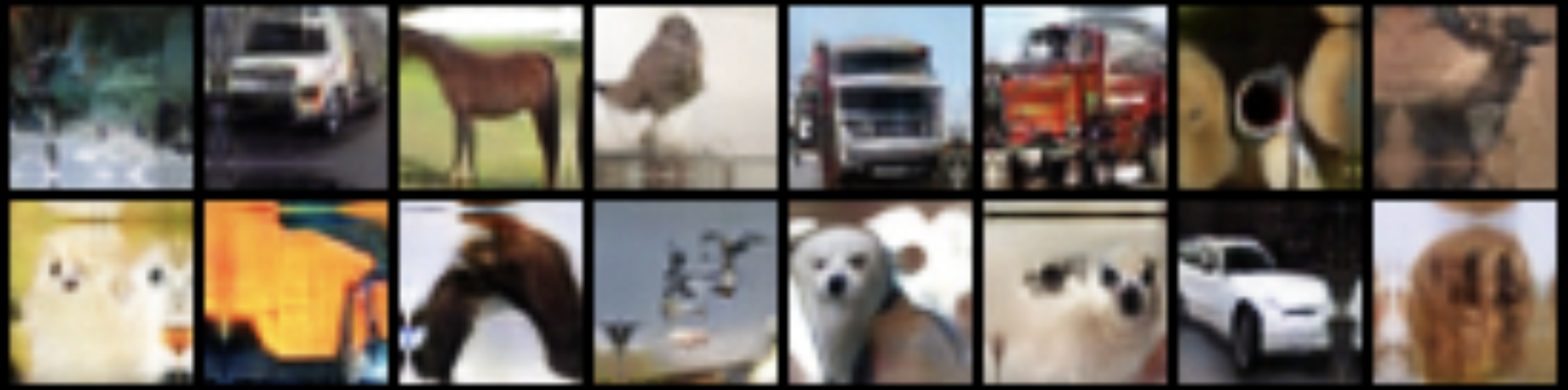}
\includegraphics[width=.44\textwidth]{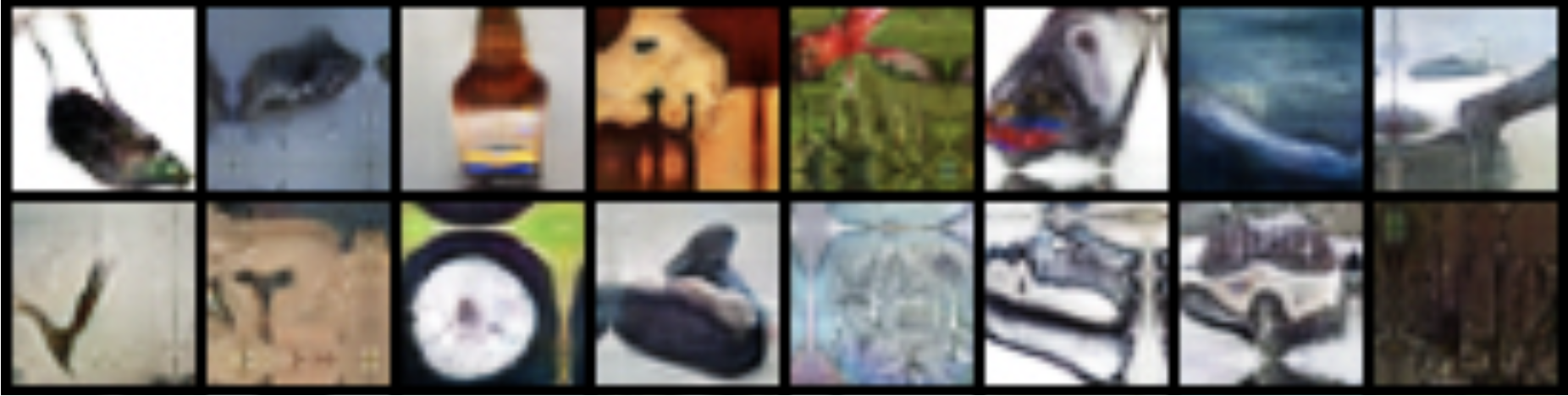}
    \vspace{-.7em}
    \caption{Unconditional samples on CIFAR10 (left) and CIFAR100 (right).}
    \vspace{-1.7em}
    \label{fig:jem_samp_main}
\end{figure}

Unconditional samples can be seen from our CIFAR10 and CIFAR100 models in Figure~\ref{fig:jem_samp_main}. Samples are refined through a simple iterative procedure using the latent space of our generator, explained in Appendix~\ref{app:mala}. Additional conditional samples can be found in Appendix~\ref{app:samp}

\paragraph{Out-of-Distribution Detection} JEM is a powerful approach for out-of-distribution detection (OOD), greatly outperforming tractable likelihood models like VAEs and flows~\citep{nalisnick2018deep}. In Table~\ref{tab:ood}, reporting AUROC \citep{hendrycks2016aurocbaseline}, we see that for all but 1 dataset, \methodname{} outperforms JEM with PCD training but under-performs contrastive training (HDGE). Intriguingly, \methodname{} performs poorly on CelebA. This result, along with the unreliable performance of DSM models at this task~\citep{li2019annealed} leads to questions regarding special benefits of MCMC training that are lost in our method as well as DSM. We leave this to future work.

\begin{table}[h]
    \centering
    \begin{tabular}{@{}l|llll@{}}
        \toprule
        Model & SVHN & CIFAR100 & CIFAR10-Interp & CelebA   \\ 
        \midrule
        JEM             & 0.67 & 0.67 & 0.65 & 0.75  \\
        HDGE            & 0.96 & 0.91 & 0.82 & 0.80  \\
        GLOW            & 0.05 & 0.51 & 0.55 & 0.57\\
        \methodname{}   & 0.83 & 0.73 & 0.86 & 0.33  \\
        \bottomrule
    \end{tabular}
    \vspace{-.7em}
    \caption{Out-of-distribution Detection. Model trained on CIFAR10. Values are AUROC ($\uparrow$).}
    \vspace{-1.0em}
    \label{tab:ood}
\end{table}

\subsection{Tabular Data}
\label{sec:ssl}
Training with \methodname{} is much more stable and easy to apply to domains beyond images where EBM training has been extensively tuned. To demonstrate this we show that JEM models trained with \methodname{} can provide a benefit to semi-supervised classification on datasets from a variety of domains. Considerable progress has been made in semi-supervised learning but the most impressive results require considerable domain knowledge~\citep{chen2020simple}. In domains like images, text, and audio such knowledge exists but for data from particle accelerators, gyroscopes, and satellites, such intuition may not be available and these techniques cannot be applied. In these settings there are far fewer options for semi-supervised learning.

We present \methodname{} as one such option. We train semi-supervised JEM models on data from a variety of continuous domains. We perform no data augmentation beyond removing redundant features and standardizing the remaining features. To further demonstrate the versatility of \methodname{} we use an identical network for each dataset and method. Full experimental details can be found in Appendix \ref{app:ssl}.

In Table \ref{tab:ssl_acc}, we find on each dataset tested, we find that \methodname{} outperforms the supervised baseline and outperforms VAT which is the strongest domain agnostic semi-supervised learning method we are aware of.


\begin{table}[h]
    \vspace{-.8em}
    \centering
    \begin{tabular}{@{}l|llll@{}}
        \toprule
        Model                   & HEPMASS       & CROP          & HUMAN         & MNIST\footnotemark   \\ 
        \midrule
        Supervised Baseline     & 81.2          & 87.8          & 81.0          & 91.0 \\
        VAT                     & 86.7          & 94.5          & 84.0          & 98.6 \\
        MEG                     & 72.1          & 87.5          & 83.5          & 94.7 \\
        JEM                     & 54.2          & 18.5          & 77.2          & 10.2 \\
        \methodname{}           & 88.3 & 94.9 & 89.9 & 98.6 \\
        \midrule
        Full-label              & 90.9          & 99.7          & 98.0          & 99.5 \\
        \bottomrule
    \end{tabular}
    \vspace{-.6em}
    \caption{Accuracy of semi-supervised learning on tabular data with 10 labeled examples per class.}
    \label{tab:ssl_acc}
\end{table}
\footnotetext{We treat MNIST as a tabular dataset since we do not use convolutional architectures.}


\vspace{-.2in}
\section{Related Work}
\vspace{-.1in}

\citet{kumar2019maximum} train EBMs using entropy-regularized generators, attempting to optimize the same objective as our own. The key difference is how the generator's entropy is regularized. \citet{kumar2019maximum} utilize a Mutual Information estimator to approximate the generator's entropy whereas we approximate the gradients of the entropy directly. The method of \citet{kumar2019maximum} requires the training of an additional MI-estimation network, but our approach only requires the optimization of the posterior variance which has considerably fewer parameters. As demonstrated in Section \ref{sec:likelihood}, their approach does not perform as well as \methodname{} for training NICE models and their generator collapses to a single point. This is likely due to the notorious difficulty of estimating MI in high dimensions and the unreliability of current approaches for this task~\citep{mcallester2020formal, song2019understanding}.

\citet{dai2019exponential} train EBMs using the same objective as \methodname{}. The key difference here is in the architecture of the generator. \citet{dai2019exponential} use generators with a restricted form, inspired by various MCMC sampling methods. In this setting, a convenient estimator for the generator's entropy can be derived. In contrast, our generators have an unconstrained architecture and we focus on entropy regularization in the unconstrained setting.

\citet{abbasnejad2019generative} train EBMs and generators as well to minimize the reverse KL-divergence. Their method differs from ours in the architecture of the generator and the method for encouraging high entropy. Their generator defines an implicit density (unlike ours which defines a latent-variable model). The entropy is maximized using a series approximation to the generator function's Jacobian log-determinant which approximates the change-of-variables for injective functions. 

\citet{gao2020flow} train EBMs using Noise Contrastive Estimation where the noise distribution is a normalizing flow. Their training objective differs from ours and their generator is restricted to having a normalizing flow architecture. These architectures do not scale as well as the GAN-style architectures we use to large image datasets.

As well there exist CoopNets~\citep{xie2018cooperative} which cooperatively train an EBM and a generator network. Architecturally, they are similar to VERA but are trained quit differently. In CoopNets, the generator is trained via maximum likelihood on its own samples refined using MCMC on the EBM. This maximum likleihood step requires MCMC as well to generate posterior samples as in \citet{pang2020learning}. In contrast, the generator in VERA is trained to minimize the reverse KL-divergence. Our method requires no MCMC and was specifically developed to alleviate the difficulties of MCMC sampling.

The estimator of \cite{dieng2019prescribed} was very influential to our work. Their work focused on applications to GANs. Our estimator could easily be applied in this setting and to implicit variational inference~\citep{titsias2019unbiased} as well but we leave this for future work. 

\vspace{-.1in}
\section{Conclusion}
\vspace{-.1in}

In this work we have presented \methodname{}, a simple and easy-to-tune approach for training unnormalized density models. We have demonstrated our approach learns high quality energy functions and models with high likelihood (when available for evaluation). We have further demonstrated the superior stability and speed of \methodname{} compared to PCD training, enabling much faster training of JEM~\citep{grathwohl2019your} while retaining the performance of the original work. We have shown that \methodname{} can train models from multiple data domains with no additional tuning. This enables the applications of JEM to semi-supervised classification on tabular data -- outperforming a strong baseline method for this task and greatly outperforming JEM with PCD training.  

\bibliography{iclr2021_conference}

\begin{thebibliography}{59}
\providecommand{\natexlab}[1]{#1}
\providecommand{\url}[1]{\texttt{#1}}
\expandafter\ifx\csname urlstyle\endcsname\relax
  \providecommand{\doi}[1]{doi: #1}\else
  \providecommand{\doi}{doi: \begingroup \urlstyle{rm}\Url}\fi

\bibitem[Abbasnejad et~al.(2019)Abbasnejad, Shi, Hengel, and
  Liu]{abbasnejad2019generative}
M~Ehsan Abbasnejad, Qinfeng Shi, Anton van~den Hengel, and Lingqiao Liu.
\newblock A generative adversarial density estimator.
\newblock In \emph{Proceedings of the IEEE Conference on Computer Vision and
  Pattern Recognition}, pp.\  10782--10791, 2019.

\bibitem[Besag(1994)]{besag1994comments}
JE~Besag.
\newblock Comments on “representations of knowledge in complex systems” by
  u. grenander and mi miller.
\newblock \emph{J. Roy. Statist. Soc. Ser. B}, 56:\penalty0 591--592, 1994.

\bibitem[Ceylan \& Gutmann(2018)Ceylan and Gutmann]{ceylan2018conditional}
Ciwan Ceylan and Michael~U Gutmann.
\newblock Conditional noise-contrastive estimation of unnormalised models.
\newblock \emph{arXiv preprint arXiv:1806.03664}, 2018.

\bibitem[Che et~al.(2020)Che, Zhang, Sohl-Dickstein, Larochelle, Paull, Cao,
  and Bengio]{che2020your}
Tong Che, Ruixiang Zhang, Jascha Sohl-Dickstein, Hugo Larochelle, Liam Paull,
  Yuan Cao, and Yoshua Bengio.
\newblock Your gan is secretly an energy-based model and you should use
  discriminator driven latent sampling.
\newblock \emph{arXiv preprint arXiv:2003.06060}, 2020.

\bibitem[Chen et~al.(2020)Chen, Kornblith, Norouzi, and Hinton]{chen2020simple}
Ting Chen, Simon Kornblith, Mohammad Norouzi, and Geoffrey Hinton.
\newblock A simple framework for contrastive learning of visual
  representations.
\newblock \emph{arXiv preprint arXiv:2002.05709}, 2020.

\bibitem[Dai et~al.(2019)Dai, Liu, Dai, He, Gretton, Song, and
  Schuurmans]{dai2019exponential}
Bo~Dai, Zhen Liu, Hanjun Dai, Niao He, Arthur Gretton, Le~Song, and Dale
  Schuurmans.
\newblock Exponential family estimation via adversarial dynamics embedding.
\newblock In \emph{Advances in Neural Information Processing Systems}, pp.\
  10979--10990, 2019.

\bibitem[Dai et~al.(2017)Dai, Almahairi, Bachman, Hovy, and
  Courville]{dai2017calibrating}
Zihang Dai, Amjad Almahairi, Philip Bachman, Eduard Hovy, and Aaron Courville.
\newblock Calibrating energy-based generative adversarial networks.
\newblock \emph{arXiv preprint arXiv:1702.01691}, 2017.

\bibitem[Dieng et~al.(2019)Dieng, Ruiz, Blei, and Titsias]{dieng2019prescribed}
Adji~B Dieng, Francisco~JR Ruiz, David~M Blei, and Michalis~K Titsias.
\newblock Prescribed generative adversarial networks.
\newblock \emph{arXiv preprint arXiv:1910.04302}, 2019.

\bibitem[Dinh et~al.(2014)Dinh, Krueger, and Bengio]{dinh2014nice}
Laurent Dinh, David Krueger, and Yoshua Bengio.
\newblock Nice: Non-linear independent components estimation.
\newblock \emph{arXiv preprint arXiv:1410.8516}, 2014.

\bibitem[Du \& Mordatch(2019)Du and Mordatch]{du2019implicit}
Yilun Du and Igor Mordatch.
\newblock Implicit generation and generalization in energy-based models.
\newblock \emph{arXiv preprint arXiv:1903.08689}, 2019.

\bibitem[Du et~al.(2020)Du, Meier, Ma, Fergus, and Rives]{du2020energy}
Yilun Du, Joshua Meier, Jerry Ma, Rob Fergus, and Alexander Rives.
\newblock Energy-based models for atomic-resolution protein conformations.
\newblock \emph{arXiv preprint arXiv:2004.13167}, 2020.

\bibitem[Gao et~al.(2020)Gao, Nijkamp, Kingma, Xu, Dai, and Wu]{gao2020flow}
Ruiqi Gao, Erik Nijkamp, Diederik~P Kingma, Zhen Xu, Andrew~M Dai, and
  Ying~Nian Wu.
\newblock Flow contrastive estimation of energy-based models.
\newblock In \emph{Proceedings of the IEEE/CVF Conference on Computer Vision
  and Pattern Recognition}, pp.\  7518--7528, 2020.

\bibitem[Grathwohl et~al.(2019)Grathwohl, Wang, Jacobsen, Duvenaud, Norouzi,
  and Swersky]{grathwohl2019your}
Will Grathwohl, Kuan-Chieh Wang, J{\"o}rn-Henrik Jacobsen, David Duvenaud,
  Mohammad Norouzi, and Kevin Swersky.
\newblock Your classifier is secretly an energy based model and you should
  treat it like one.
\newblock \emph{arXiv preprint arXiv:1912.03263}, 2019.

\bibitem[Grathwohl et~al.(2020)Grathwohl, Wang, Jacobsen, Duvenaud, and
  Zemel]{grathwohllearning}
Will Grathwohl, Kuan-Chieh Wang, J{\"o}rn-Henrik Jacobsen, David Duvenaud, and
  Richard Zemel.
\newblock Learning the stein discrepancy for training and evaluating
  energy-based models without sampling.
\newblock 2020.

\bibitem[Gutmann \& Hyv{\"a}rinen(2010)Gutmann and
  Hyv{\"a}rinen]{gutmann2010noise}
Michael Gutmann and Aapo Hyv{\"a}rinen.
\newblock Noise-contrastive estimation: A new estimation principle for
  unnormalized statistical models.
\newblock In \emph{Proceedings of the Thirteenth International Conference on
  Artificial Intelligence and Statistics}, pp.\  297--304, 2010.

\bibitem[Hendrycks \& Gimpel(2016)Hendrycks and
  Gimpel]{hendrycks2016aurocbaseline}
Dan Hendrycks and Kevin Gimpel.
\newblock A baseline for detecting misclassified and out-of-distribution
  examples in neural networks.
\newblock \emph{arXiv preprint arXiv:1610.02136}, 2016.

\bibitem[Heusel et~al.(2017)Heusel, Ramsauer, Unterthiner, Nessler, and
  Hochreiter]{heusel2017gans}
Martin Heusel, Hubert Ramsauer, Thomas Unterthiner, Bernhard Nessler, and Sepp
  Hochreiter.
\newblock Gans trained by a two time-scale update rule converge to a local nash
  equilibrium.
\newblock In \emph{Advances in neural information processing systems}, pp.\
  6626--6637, 2017.

\bibitem[Hill et~al.(2020)Hill, Mitchell, and Zhu]{hill2020stochastic}
Mitch Hill, Jonathan Mitchell, and Song-Chun Zhu.
\newblock Stochastic security: Adversarial defense using long-run dynamics of
  energy-based models.
\newblock \emph{arXiv preprint arXiv:2005.13525}, 2020.

\bibitem[Hyv{\"a}rinen(2005)]{hyvarinen2005estimation}
Aapo Hyv{\"a}rinen.
\newblock Estimation of non-normalized statistical models by score matching.
\newblock \emph{Journal of Machine Learning Research}, 6\penalty0
  (Apr):\penalty0 695--709, 2005.

\bibitem[Ingraham et~al.(2019)Ingraham, Riesselman, Sander, and
  Marks]{ingraham2019learning}
John Ingraham, Adam~J Riesselman, Chris Sander, and Debora~S Marks.
\newblock Learning protein structure with a differentiable simulator.
\newblock In \emph{ICLR}, 2019.

\bibitem[Ioffe \& Szegedy(2015)Ioffe and Szegedy]{ioffe2015batch}
Sergey Ioffe and Christian Szegedy.
\newblock Batch normalization: Accelerating deep network training by reducing
  internal covariate shift.
\newblock \emph{arXiv preprint arXiv:1502.03167}, 2015.

\bibitem[Kingma \& Ba(2014)Kingma and Ba]{kingma2014adam}
Diederik~P Kingma and Jimmy Ba.
\newblock Adam: A method for stochastic optimization.
\newblock \emph{arXiv preprint arXiv:1412.6980}, 2014.

\bibitem[Kingma \& Welling(2013)Kingma and Welling]{kingma2013auto}
Diederik~P Kingma and Max Welling.
\newblock Auto-encoding variational bayes.
\newblock \emph{arXiv preprint arXiv:1312.6114}, 2013.

\bibitem[Kingma \& Dhariwal(2018)Kingma and Dhariwal]{kingma2018glow}
Durk~P Kingma and Prafulla Dhariwal.
\newblock Glow: Generative flow with invertible 1x1 convolutions.
\newblock In \emph{Advances in neural information processing systems}, pp.\
  10215--10224, 2018.

\bibitem[Kong(1992)]{kong1992note}
Augustine Kong.
\newblock A note on importance sampling using standardized weights.
\newblock \emph{University of Chicago, Dept. of Statistics, Tech. Rep}, 348,
  1992.

\bibitem[Kumar et~al.(2019)Kumar, Ozair, Goyal, Courville, and
  Bengio]{kumar2019maximum}
Rithesh Kumar, Sherjil Ozair, Anirudh Goyal, Aaron Courville, and Yoshua
  Bengio.
\newblock Maximum entropy generators for energy-based models.
\newblock \emph{arXiv preprint arXiv:1901.08508}, 2019.

\bibitem[Li et~al.(2019)Li, Chen, and Sommer]{li2019annealed}
Zengyi Li, Yubei Chen, and Friedrich~T Sommer.
\newblock Annealed denoising score matching: Learning energy-based models in
  high-dimensional spaces.
\newblock \emph{arXiv preprint arXiv:1910.07762}, 2019.

\bibitem[Lim et~al.(2020)Lim, Courville, Pal, and Huang]{lim2020ar}
Jae~Hyun Lim, Aaron Courville, Christopher Pal, and Chin-Wei Huang.
\newblock Ar-dae: Towards unbiased neural entropy gradient estimation.
\newblock \emph{arXiv preprint arXiv:2006.05164}, 2020.

\bibitem[Liu \& Abbeel(2020)Liu and Abbeel]{liu2020hybrid}
Hao Liu and Pieter Abbeel.
\newblock Hybrid discriminative-generative training via contrastive learning.
\newblock \emph{arXiv preprint arXiv:2007.09070}, 2020.

\bibitem[Martens et~al.(2012)Martens, Sutskever, and
  Swersky]{martens2012estimating}
James Martens, Ilya Sutskever, and Kevin Swersky.
\newblock Estimating the hessian by back-propagating curvature.
\newblock \emph{arXiv preprint arXiv:1206.6464}, 2012.

\bibitem[McAllester \& Stratos(2020)McAllester and
  Stratos]{mcallester2020formal}
David McAllester and Karl Stratos.
\newblock Formal limitations on the measurement of mutual information.
\newblock In \emph{International Conference on Artificial Intelligence and
  Statistics}, pp.\  875--884, 2020.

\bibitem[Miyato et~al.(2018)Miyato, Kataoka, Koyama, and
  Yoshida]{miyato2018spectral}
Takeru Miyato, Toshiki Kataoka, Masanori Koyama, and Yuichi Yoshida.
\newblock Spectral normalization for generative adversarial networks.
\newblock \emph{arXiv preprint arXiv:1802.05957}, 2018.

\bibitem[Nalisnick et~al.(2018)Nalisnick, Matsukawa, Teh, Gorur, and
  Lakshminarayanan]{nalisnick2018deep}
Eric Nalisnick, Akihiro Matsukawa, Yee~Whye Teh, Dilan Gorur, and Balaji
  Lakshminarayanan.
\newblock Do deep generative models know what they don't know?
\newblock \emph{arXiv preprint arXiv:1810.09136}, 2018.

\bibitem[Neal et~al.(2011)]{neal2011mcmc}
Radford~M Neal et~al.
\newblock Mcmc using hamiltonian dynamics.
\newblock \emph{Handbook of markov chain monte carlo}, 2\penalty0
  (11):\penalty0 2, 2011.

\bibitem[Nijkamp et~al.(2019{\natexlab{a}})Nijkamp, Hill, Han, Zhu, and
  Wu]{nijkamp2019anatomy}
Erik Nijkamp, Mitch Hill, Tian Han, Song-Chun Zhu, and Ying~Nian Wu.
\newblock On the anatomy of mcmc-based maximum likelihood learning of
  energy-based models.
\newblock \emph{arXiv preprint arXiv:1903.12370}, 2019{\natexlab{a}}.

\bibitem[Nijkamp et~al.(2019{\natexlab{b}})Nijkamp, Hill, Zhu, and
  Wu]{nijkamp2019learning}
Erik Nijkamp, Mitch Hill, Song-Chun Zhu, and Ying~Nian Wu.
\newblock Learning non-convergent non-persistent short-run mcmc toward
  energy-based model.
\newblock In \emph{Advances in Neural Information Processing Systems}, pp.\
  5232--5242, 2019{\natexlab{b}}.

\bibitem[No{\'e} et~al.(2019)No{\'e}, Olsson, K{\"o}hler, and
  Wu]{noe2019boltzmann}
Frank No{\'e}, Simon Olsson, Jonas K{\"o}hler, and Hao Wu.
\newblock Boltzmann generators: Sampling equilibrium states of many-body
  systems with deep learning.
\newblock \emph{Science}, 365\penalty0 (6457):\penalty0 eaaw1147, 2019.

\bibitem[Pang et~al.(2020{\natexlab{a}})Pang, Han, Nijkamp, Zhu, and
  Wu]{pang2020learning}
Bo~Pang, Tian Han, Erik Nijkamp, Song-Chun Zhu, and Ying~Nian Wu.
\newblock Learning latent space energy-based prior model.
\newblock \emph{Advances in Neural Information Processing Systems}, 33,
  2020{\natexlab{a}}.

\bibitem[Pang et~al.(2020{\natexlab{b}})Pang, Xu, Li, Song, Ermon, and
  Zhu]{pang2020efficient}
Tianyu Pang, Kun Xu, Chongxuan Li, Yang Song, Stefano Ermon, and Jun Zhu.
\newblock Efficient learning of generative models via finite-difference score
  matching.
\newblock \emph{arXiv preprint arXiv:2007.03317}, 2020{\natexlab{b}}.

\bibitem[Radford et~al.(2015)Radford, Metz, and
  Chintala]{radford2015unsupervised}
Alec Radford, Luke Metz, and Soumith Chintala.
\newblock Unsupervised representation learning with deep convolutional
  generative adversarial networks.
\newblock \emph{arXiv preprint arXiv:1511.06434}, 2015.

\bibitem[Ranganath et~al.(2016)Ranganath, Tran, and
  Blei]{ranganath2016hierarchical}
Rajesh Ranganath, Dustin Tran, and David Blei.
\newblock Hierarchical variational models.
\newblock In \emph{International Conference on Machine Learning}, pp.\
  324--333, 2016.

\bibitem[Rezende \& Mohamed(2015)Rezende and Mohamed]{rezende2015variational}
Danilo~Jimenez Rezende and Shakir Mohamed.
\newblock Variational inference with normalizing flows.
\newblock \emph{arXiv preprint arXiv:1505.05770}, 2015.

\bibitem[Rhodes et~al.(2020)Rhodes, Xu, and Gutmann]{rhodes2020telescoping}
Benjamin Rhodes, Kai Xu, and Michael~U Gutmann.
\newblock Telescoping density-ratio estimation.
\newblock \emph{arXiv preprint arXiv:2006.12204}, 2020.

\bibitem[Salimans et~al.(2016)Salimans, Goodfellow, Zaremba, Cheung, Radford,
  and Chen]{salimans2016improved}
Tim Salimans, Ian Goodfellow, Wojciech Zaremba, Vicki Cheung, Alec Radford, and
  Xi~Chen.
\newblock Improved techniques for training gans.
\newblock In \emph{Advances in neural information processing systems}, pp.\
  2234--2242, 2016.

\bibitem[Singh \& P{\'o}czos(2016)Singh and P{\'o}czos]{singh2016analysis}
Shashank Singh and Barnab{\'a}s P{\'o}czos.
\newblock Analysis of k-nearest neighbor distances with application to entropy
  estimation.
\newblock \emph{arXiv preprint arXiv:1603.08578}, 2016.

\bibitem[Song \& Ermon(2019{\natexlab{a}})Song and
  Ermon]{song2019understanding}
Jiaming Song and Stefano Ermon.
\newblock Understanding the limitations of variational mutual information
  estimators.
\newblock \emph{arXiv preprint arXiv:1910.06222}, 2019{\natexlab{a}}.

\bibitem[Song \& Ermon(2019{\natexlab{b}})Song and Ermon]{song2019generative}
Yang Song and Stefano Ermon.
\newblock Generative modeling by estimating gradients of the data distribution.
\newblock In \emph{Advances in Neural Information Processing Systems}, pp.\
  11918--11930, 2019{\natexlab{b}}.

\bibitem[Song \& Ermon(2020)Song and Ermon]{song2020improved}
Yang Song and Stefano Ermon.
\newblock Improved techniques for training score-based generative models.
\newblock \emph{arXiv preprint arXiv:2006.09011}, 2020.

\bibitem[Song et~al.(2020)Song, Garg, Shi, and Ermon]{song2020sliced}
Yang Song, Sahaj Garg, Jiaxin Shi, and Stefano Ermon.
\newblock Sliced score matching: A scalable approach to density and score
  estimation.
\newblock In \emph{Uncertainty in Artificial Intelligence}, pp.\  574--584.
  PMLR, 2020.

\bibitem[Song \& Ou(2018)Song and Ou]{song2018learning}
Yunfu Song and Zhijian Ou.
\newblock Learning neural random fields with inclusive auxiliary generators.
\newblock \emph{arXiv preprint arXiv:1806.00271}, 2018.

\bibitem[Tieleman(2008)]{tieleman2008training}
Tijmen Tieleman.
\newblock Training restricted boltzmann machines using approximations to the
  likelihood gradient.
\newblock In \emph{International Conference on Machine Learning}, pp.\
  1064--1071, 2008.

\bibitem[Tipping \& Bishop(1999)Tipping and Bishop]{tipping1999probabilistic}
Michael~E Tipping and Christopher~M Bishop.
\newblock Probabilistic principal component analysis.
\newblock \emph{Journal of the Royal Statistical Society: Series B (Statistical
  Methodology)}, 61\penalty0 (3):\penalty0 611--622, 1999.

\bibitem[Titsias \& Ruiz(2019)Titsias and Ruiz]{titsias2019unbiased}
Michalis~K Titsias and Francisco Ruiz.
\newblock Unbiased implicit variational inference.
\newblock In \emph{The 22nd International Conference on Artificial Intelligence
  and Statistics}, pp.\  167--176, 2019.

\bibitem[Vincent(2011)]{vincent2011connection}
Pascal Vincent.
\newblock A connection between score matching and denoising autoencoders.
\newblock \emph{Neural computation}, 23\penalty0 (7):\penalty0 1661--1674,
  2011.

\bibitem[Wainwright \& Jordan(2008)Wainwright and
  Jordan]{wainwright2008graphical}
Martin~J Wainwright and Michael~Irwin Jordan.
\newblock \emph{Graphical models, exponential families, and variational
  inference}.
\newblock Now Publishers Inc, 2008.

\bibitem[Welling \& Teh(2011)Welling and Teh]{welling2011bayesian}
Max Welling and Yee~W Teh.
\newblock Bayesian learning via stochastic gradient langevin dynamics.
\newblock In \emph{Proceedings of the 28th international conference on machine
  learning (ICML-11)}, pp.\  681--688, 2011.

\bibitem[Xie et~al.(2018)Xie, Lu, Gao, Zhu, and Wu]{xie2018cooperative}
Jianwen Xie, Yang Lu, Ruiqi Gao, Song-Chun Zhu, and Ying~Nian Wu.
\newblock Cooperative training of descriptor and generator networks.
\newblock \emph{IEEE transactions on pattern analysis and machine
  intelligence}, 42\penalty0 (1):\penalty0 27--45, 2018.

\bibitem[Zagoruyko \& Komodakis(2016)Zagoruyko and
  Komodakis]{zagoruyko2016wide}
Sergey Zagoruyko and Nikos Komodakis.
\newblock Wide residual networks.
\newblock \emph{arXiv preprint arXiv:1605.07146}, 2016.

\bibitem[Zhao et~al.()Zhao, Jacobsen, and Grathwohl]{zhaojoint}
Stephen Zhao, J{\"o}rn-Henrik Jacobsen, and Will Grathwohl.
\newblock Joint energy-based models for semi-supervised classification.

\end{thebibliography}
\bibliographystyle{iclr2021_conference}

\newpage

\appendix

\section{Key Derivations}
\subsection{Derivation of Variational Log-Partition Function}
\label{app:dual_der}
Here we show that the variational optimization given in \eqref{eq:fencheldual} recovers $\log Z(\theta)$.

\begin{align*}
    & \max_{q} \E_{q(x)}[f_\theta(x)] + H(q) \\
    &= \max_{q} \int_x q(x) f_\theta(x) \mathrm{d}x - \int_x q(x) \log(q(x)) \mathrm{d}x \\
    &= \max_{q} \int_x q(x) \log\left(\frac{\exp(f_\theta(x))}{q(x)}\right) \mathrm{d}x \\
    &= \max_{q} \int_x q(x) \log\left(\frac{\exp(f_\theta(x))}{q(x)}\right) \mathrm{d}x - \log Z(\theta) + \log Z(\theta) \\
    &= \max_{q} \int_x q(x) \log\left(\frac{\exp(f_\theta(x)) / Z(\theta)}{q(x)}\right) \mathrm{d}x  + \log Z(\theta) \\
    &= \max_{q} -\mathrm{KL}(q(x) \| p_\theta(x)) + \log Z(\theta) \\
    &= \log Z(\theta).
\end{align*}

\subsection{Score Function Estimator}
\label{app:score_fn}
Here we derive the equivalent expression for $\nabla_x \log q_\phi (x)$ given in Equation \ref{eq:grad_logqx}.

\begin{align*}
    \nabla_x \log q_\phi (x) &= \frac{\nabla_x q_\phi(x)}{q_\phi(x)} \\
    &=  \frac{\nabla_x \int_z q_\phi(x, z) dz}{q_\phi(x)} \\
    &=  \frac{\int_z \nabla_x q_\phi(x, z) dz}{q_\phi(x)} \\
    &=  \int_z\frac{\nabla_x q_\phi(x, z) }{q_\phi(x)}dz \\
    &= \int_z \frac{\nabla_x q_\phi(x \mid z) q_\phi(z)}{q_\phi(x)} dz \\
    &= \int_z \frac{(\nabla_x \log q_\phi(x \mid z)) q_\phi(x \mid z) q_\phi(z)}{q_\phi(x)} dz \\
    &= \E_{q_\phi(z|x)}[\nabla_x \log q_\phi(x|z)].
\end{align*}

\subsection{Entropy Gradient Estimator}
\label{app:ent_est}
From Equation \ref{eq:ent_est} we have
\begin{align}
    \nabla_\phi H(q_\phi) &=  \E_{p(z_0)p(\epsilon)}[\nabla_x \log q_\phi (x)^T \nabla_\phi x(z_0, \epsilon)].\nonumber
\end{align}
Plugging in our score function estimator gives
\begin{align}
    \nabla_\phi H(q_\phi) &=\E_{p(z_0)p(\epsilon)}[\nabla_x \log q_\phi (x)^T \nabla_\phi x(z_0, \epsilon)]\\
    &= \E_{p(z_0)p(\epsilon)}\left[\E_{q_\phi(z\mid x)}\left[ \nabla_x \log q_\phi(x \mid z) \right]^T \nabla_\phi x(z_0, \epsilon)\right] \nonumber\\
    &= \E_{p(z_0)p(\epsilon)}\left[\E_{\xi_\eta(z\mid z_0)}\left[ \frac{q_\phi(z\mid x)}{\xi_\eta(z\mid z_0)}\nabla_x \log q_\phi(x \mid z) \right]^T \nabla_\phi x(z_0, \epsilon)\right]\nonumber\\
    &= \E_{p(z_0)p(\epsilon)}\left[\E_{\xi_\eta(z\mid z_0)}\left[ \frac{q_\phi(x, z)}{q_\phi(x)\xi_\eta(z\mid z_0)}\nabla_x \log q_\phi(x \mid z) \right]^T \nabla_\phi x(z_0, \epsilon)\right]\nonumber\\
    &\approx  \E_{p(z_0)p(\epsilon)}\left[\left[\sum_{i=1}^{k} \frac{w_i}{\sum_{j=1}^{k} w_j} \nabla_x \log q_\phi(x \mid z_i)\right]^T \nabla_\phi x(z_0, \epsilon)\right]\nonumber
    \label{eq:ent_est_long}
\end{align}
where $\{z_i\}_{i=1}^k \sim \xi(z\mid z_0)$ and $w_i \equiv \frac{q_\phi(z_i, x)}{\xi(z_i \mid z_0)}$.

\subsubsection{Discussion}
We discuss when the approximations in Equation \ref{eq:ent_est_long} will hold. The importance sampling estimator will be biased when $q_\psi(z \mid x)$ differs greatly from $\xi(z \mid z_0)$. Since the generator function $g_\psi(z)$ is a smooth, Lipschitz function (as are most neural networks) and the output Gaussian noise is small, the space of $z$ values which could have generated $x$ should be concentrated near $z_0$. In these settings, $z_0$ should be useful for predicting $q(z|x)$.

The accuracy of this approximation is based on the dimension of $z$ compared to $x$ and the Lipschitz constant of $g_\psi$. In all settings we tested, dim$(z) \ll $ dim$(x)$ where this approximation should hold. If dim$(z) \gg $ dim$(x)$ the curse of dimensionality would take effect and $z_0$ would be less and less informative about $q(z|x)$. In settings such as these, we do not believe our approach would be as effective. Thankfully, almost all generator architectures we are aware of have dim$(z) \ll $ dim$(x)$. The approximation could also break down if the Lipschitz constant blew up. We find this does not happen in practice, but this can be addressed with many forms of regularization and normalization.


\newpage
\section{Experimental Details}

\subsection{Hyperparameter Recommendations}

\begin{table}[h]
    \centering
    \begin{tabular}{@{}l|l|l}
        \toprule
        Type & Hyperparameter & Value \\
        \midrule
        Optimization    & $\beta_1$ (ADAM) & $0$  \\
                        & $\beta_2$ (ADAM) & $0.9$  \\
                        & learning rate (energy) & ${10^{-4}}^{*}$\\
                        & learning rate (generator) & $2 \cdot {10^{-4}}^*$ \\
                        & learning rate (posterior) & $2 \cdot {10^{-4}}^*$ \\
        \midrule
        Regularization  &  $\lambda$ (entropy regularization weight) & ${10^{-4}}^{\dagger}$\\
                        &  $\gamma$ (gradient norm penalty) & $0.1$  \\
        \midrule
        JEM             &  $\alpha$ (classification weight)& $\{1, 10, 30, 100\}^*$ \\
                        &  $\beta$ (classification entropy)& $\{1, 0.1 * \alpha, \alpha\}$ \\
        \bottomrule
    \end{tabular}
    \caption{Hyperparameters for \methodname{}.}
    \label{tab:general_hyperparams}
\end{table}

*When $\alpha > 1$, learning rates were divided by $\alpha$.

$^\dagger$We found $\lambda=10^{-4}$ to work best on large image datasets, but in general we recommend starting with $\lambda=1$ and trying successively smaller values of $\lambda$ until training is stable.

We give some general tips on how to set hyperparameters when training \methodname{} in Table \ref{tab:general_hyperparams}. In all \methodname{} experiments, we use the gradient norm penalty with weight $0.1$. This was not tuned during our experiments. When using \methodname{} and MEG we train with Adam~\citep{kingma2014adam} and set $\beta_1 = 0$, $\beta_2 = .9$ as is standard in the GAN literature~\citep{miyato2018spectral}. In general, we recommend setting the learning rate for the generator to twice the learning rate of the energy function and equal to the learning rate of the approximate posterior sampler.

\subsubsection{Impact of $\lambda$}
Let us rewrite the generator's training objective
\begin{align}
    \mathcal{L}(q; \lambda) = E_{q(x)}[f_\theta(x)] + \lambda H(q).
\end{align}
We can easily see that this objective is equivalent (up to a multiplicative constant) to
\begin{align}
    E_{q(x)}\left[\frac{f_\theta(x)}{\lambda}\right] + H(q).
    \label{eq:tempered}
\end{align}
From this, it is clear that maximizing Equation \ref{eq:tempered} is the same as minimizing the KL-divergence between $q$ and a tempered version of $p_\theta(x)$ defined as 
\begin{align}
    \frac{e^{f_\theta(x)/\lambda}}{Z}.
\end{align}

Tempering like this is standard practice in EBM training and is done in many recent works. Tempering has the effect of increasing the weight of the gradient signal in SGLD sampling relative to the added Gaussian noise. In all of \citet{du2019implicit, grathwohl2019your, nijkamp2019anatomy, nijkamp2019learning} the SGLD samplers used a temperature of $1/\lambda$ = 20,000. This value is near the value of $10,000$ that we use in this work. 

Thus we can see that our most important hyper-parameter is actually the temperature of the sampler's target distribution. Ideally this temperature would be set to 1, but this can lead to unstable training (as it does with SGLD). To train high quality models, we recommend setting $\lambda$ as close to 1 as possible, decreasing if training becomes unstable.

\subsection{Toy Data Visualizations}
\label{app:toy}
We train simple energy functions on a 2D toy data. The EBM is a 2-layer MLP with 100 hidden units per layer using the Leaky-ReLU nonlinearity with negative slope $.2$. The generator is a 2-layer MLP with 100 hidden units per layer and uses batch normalization~\citep{ioffe2015batch} with ReLU nonlinearities. All models were trained for 100,000 iterations with all learning rates set to .001 and used the Adam optimizer~\citep{kingma2014adam}.

The PCD models were trained using an SGLD sampler and a replay buffer with 10,000 examples, reinitialized every iteration with $5\%$ probability. We used 20 steps of SGLD per training iteration to make runtime consistent with \methodname{}. We tested $\sigma$ values outside of the presented range but smaller values did not produce decent samples or energy functions and for larger values, training diverged.

\subsection{Training NICE Models}
\label{app:nice}
The NICE models were exactly as in \citet{song2020sliced}. They have 4 coupling layers and each coupling layer had 5 hidden layers. Each hidden layer has 1000 units and uses the Softplus non-linearity. We preprocessed the data as in \citet{song2020sliced} by scaling the data to the range $[0, 1]$, adding uniform noise in the range $[-1/512, 1/512]$, clipping to the range $[.001, .999]$ and applying the logit transform $\log (x) - \log(1 - x)$. All models were trained for 400 epochs with the Adam optimizer~\citep{kingma2014adam} with $\beta_1 = 0$ and $\beta_2 = .9$. We use a batch size of 128 for all models. We re-ran the score matching model of \citet{song2020sliced} to train for 400 epochs as well and found it did not improve performance as its best test performance happens very early in training. 

For all generator-based training methods we use the same fixed generator architecture. The generator has a latent-dimension of 100 and 2 hidden layers with 500 units each. We use the Softplus nonlinearity and batch-normalization~\citep{ioffe2015batch} as is common with generator networks.

For \methodname{} the hyper-parameters we searched over were the learning rates for the NICE model and for the generator. Compared to \citet{song2020sliced} we needed to use much lower learning rates. We searched over learning rates in $\{.0003, .00003, .000003\}$ for both the generator and energy function. We found $.000003$ to work best for the energy function and $.0003$ to work best for the generator. This makes intuitive sense since the generator needs to be fully optimized for the bound on likelihood to be tight. When equal learning rates were used $(.0003, .0003)$ we observed high sample quality from the \emph{generator} but exact samples and likelihoods from the NICE model were very poor. 

For PCD we search over learning rates in $\{.0003, .00003, .000003\}$, the number of MCMC steps in $\{20, 40\}$ and the SGLD noise standard-deviation in $\{1.0, 0.1\}$. All models with 20 steps and SGLD standard-deviation 1.0 quickly diverged. Our best model used learning rate .000003, step-size 0.1, and 40 steps. We tested the gradient-norm regularizer from \citet{kumar2019maximum} and found it decreased performance for PCD trained models. Most models with 20 MCMC steps diverged early in training. 

For review the MCMC sampler we use is stochastic gradient Langevin dynamics~\citep{welling2011bayesian}. This sampler updates its samples by
\begin{align}
    x_t = x_{t-1} + \frac{\sigma^2}{2}\nabla_x f_\theta(x) + \epsilon  \sigma, \qquad \epsilon \sim \mathcal{N}(0, I)
\end{align}
where $\sigma$ is the noise standard-deviation and is a parameter of the sampler.

For Maximum Entropy Generators (MEG)~\citep{kumar2019maximum} we must choose a mutual-information estimation network. We follow their work and use an MLP with LeakyReLU nonlinearities with negative slope $.2$. Our network mirrors the generator and has 3 hidden layers with 500 units each. We searched over the same hyper-parameters as \methodname{}. We found MEG to perform almost identically to training with no entropy regularization at all. We believe this has to do with the challenges of estimating MI in high dimensions~\citet{song2019understanding}.

For CoopNets~\citep{xie2018cooperative} we use the same flow and generator architectures as \methodname{}. Following the MNIST experiments in \citet{xie2018cooperative} we train using 10 SGLD steps per iteration. We tried the recommended learning rates of .007 and .0001 for the flow and generator, respectively but found this to lead quick divergence. For this reason, we search over learning rates for the flow and generator from $\{.0003, .00003, .000003\}$ as we did for VERA and found the best combination to be $.000003$ for the flow and $.0003$ for the generator. Other combinations resulted in higher quality generator samples but \emph{much} worse likelihood values. We tested the recommended SGLD step-size of $.002$ and found this to lead to divergence as well in this setup. Thus, we searched over larger values of $\{.002, .02, .1\}$ found $.1$ to perform the best, as with PCD.

\subsection{Estimation of Bias of Entropy Regularizer}
\label{app:bias}
\begin{figure}[h] 
\centering
\includegraphics[width=.3 \textwidth]{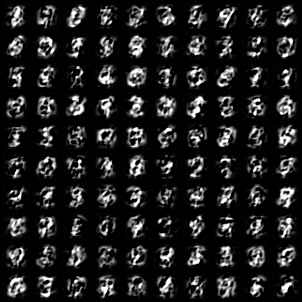}
    \caption{Samples from linear model trained with PCA on MNIST.}
    \label{fig:pca_samp}
\end{figure}

If we restrict the form of our generator to a linear function
\begin{align*}
    x = Wz + \mu + \sigma \epsilon, \qquad z,\epsilon \sim \mathcal{N}(0, I)
\end{align*}
then we have
\begin{align*}
    q(x|z) = \mathcal{N}(Wz + \mu, \sigma^2 I),\qquad q(x) = \mathcal{N}(\mu, W^T W + \sigma^2 I)
\end{align*}
meaning we can exactly compute $\log q(x)$, and $\nabla_x \log q(x)$ which is the quantity that \methodname{} (HMC) approximates with the HMC estimator from \citet{dieng2019prescribed} and we approximate with \methodname{}. To explore this, we fit a PCA model on MNIST and recover the parameters $W, \mu$ of the linear generator and the noise parameter $\sigma$. Samples from this model can be seen in Figure \ref{fig:pca_samp}. 

Both \methodname{} and \methodname{} (HMC) have some parameters which are tuned automatically to improve the estimator throughout training. For \methodname{} this is the posterior variance which is optimized according to Equation \ref{eq:iselbo} with Adam with default hyperparameters. For \methodname{} (HMC) this is the stepsize of HMC which is tuned automatically as outlined in \citep{dieng2019prescribed}. Both estimators were trained with a learning rate of $.01$ for 500 iterations with a batch size of 5000. Samples from the estimators during training were taken with the default parameters, in particular the number of burn-in steps or the number of importance samples was not varied as they were during evaluation. 

The bias of the estimators were evaluated on a batch of 10 samples from the generator. For each example in the batch, 5000 estimates were taken from the estimator and averaged to be taken as an estimate of the score function for this batch example. This estimate of the score function was subtracted from the true score function and then averaged over all dimensions and examples in the batch and taken as an estimate of the bias per dimension.

For \methodname{} (HMC) we varied the number of burn-in steps used for samples to evaluate the bias of the estimator. We also tried to increase the number of posterior samples taken off the chain, but we found that this did not clearly reduce the bias of this estimator as the number of samples increased. For \methodname{} we computed the bias on the default of 20 importance samples.

\subsection{Mode Counting}
\label{app:mode}

We train \methodname{} without labels on the \textsc{StackedMNIST} dataset. This dataset consists of 60,000 examples of 3 stacked \textsc{MNIST} images sampled with replacement from the original 60,000 \textsc{MNIST} train set.  As in \citep{dieng2019prescribed} we resize the images to $64 \times 64$ and use the DCGAN~\citep{radford2015unsupervised} architecture for both the energy-function and generator. We use a latent code of 100 dimensions. We train for 17 epochs with a learning rate of $.001$ and batch size $100$.

We estimate the number of modes captured by taking $S=10,000$ samples as in \citep{dieng2019prescribed} and classifying each digit of the 3 stacked images separately with a pre-trained classifier on MNIST.

\subsection{Effective Sample Size}
\label{app:ess}
When performing importance sampling, the quality of the proposal distribution has a large impact. If the proposal is chosen poorly, then typically 1 sample will dominate in the expectation. This can be quantified using the effective sample size~\citep{kong1992note} (ESS) which is defined as
\begin{align}
    w_i &= \frac{p(x)}{q(x)}\nonumber\\
    \tilde{w_i} &= \frac{w_i}{\sum_{j=1}^N w_j}\nonumber\\
    \text{ESS} &= \frac{1}{\sum_{i=1}^N \tilde{w_i}^2}
\end{align}
where $p(x)$ is the target distribution, $q(x)$ is the proposal distribution and $N$ is the number of samples. If the self-normalized importance weights are dominated by one weight close to 1 then the ESS will be 1. If the proposal distribution is identical to the target, so that the self-normalized importance weights are then uniform, then the ESS will be $N$. When ESS $=N$, importance sampling is as efficient as using the target distribution.

\begin{figure}[h] 
\centering
\includegraphics[width=.5 \textwidth]{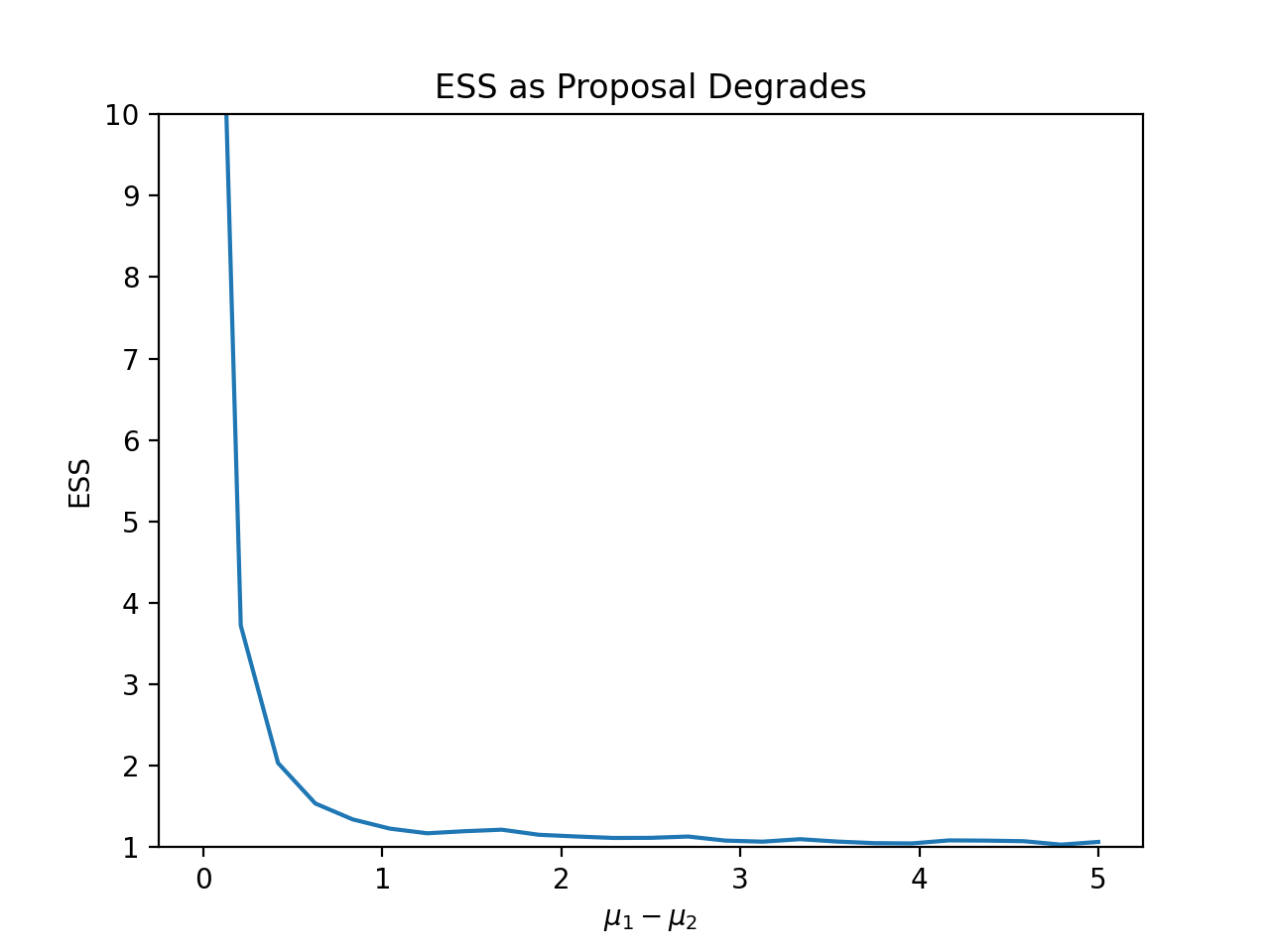}
    \caption{Effective sample size of importance sampling as the proposal degrades.}
    \label{fig:ess}
\end{figure}

To understand the effect of the proposal distribution on ESS we plot the ESS when doing importance sampling with 20 importance samples from a 128-dimensional Gaussian target distribution with $\mu = 0$ and $\Sigma = I$. We use a proposal which is a 128-dimensional Gaussian with $\mu$ increasing from $0$ to $5$. Results can be seen in Figure \ref{fig:ess}. We see when the means differ by greater than 2, the ESS is approximately 1.0 and importance sampling has effectively failed.

\subsection{JEM Models}
\label{app:jem}
Our energy function used the same architecture as in \citet{grathwohl2019your, liu2020hybrid}, a Wide ResNet~\citet{zagoruyko2016wide} 28-10 with batch-norm removed and with dropout. Our generator architecture is identical to \citet{miyato2018spectral}. As in \citet{grathwohl2019your, liu2020hybrid} we set the learning rate for the energy function equal to 0.0001. We set the learning rate for the generator equal to 0.0002. We train for 200 epochs using the Adam optimizer with $\beta_1 = 0$ and $\beta_2 = .9$. We set the batch size to 64. Results presented are from the models after 200 epochs with no early stopping. We believe better results could be obtained from further training.

We trained models with $\alpha \in \{1, 30, 100\}$ and found classification to be best with $\alpha=100$ and generation to be best with $\alpha=1$.

Prior work on PCD EBM training~\citep{grathwohl2019your, du2019implicit, nijkamp2019anatomy, nijkamp2019learning} recommends adding Gaussian noise to the data to stabilize training. Without this, PCD training of JEM models very quickly diverges. Early in our experiments we found training with \methodname{} was stable without the addition of Gaussian noise so we do not use it.

As mentioned in \citet{dieng2019prescribed}, when the strength of the entropy regularizer $\lambda$ is too high, the generator may fall into a degenerate optimum where it just outputs high-entropy Gaussian noise. To combat this, as suggested in \citet{dieng2019prescribed}, we decrease the strength of $\lambda$ to .0001 for all JEM experiments. This value was chosen by decreasing $\lambda$ from 1.0 by factors of 10 until learning took place (quantified by classification accuracy).

\subsubsection{MCMC Sample Refinement}
\label{app:mala}
\begin{figure}[h]  
\centering
\includegraphics[width=.48\textwidth]{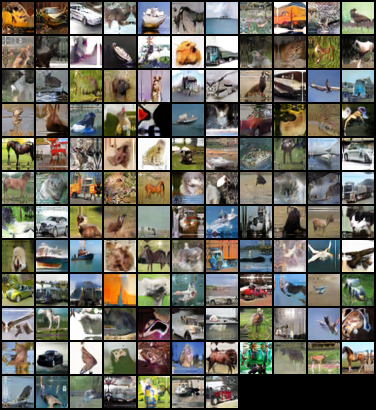}
\hspace{1em}
\includegraphics[width=.48\textwidth]{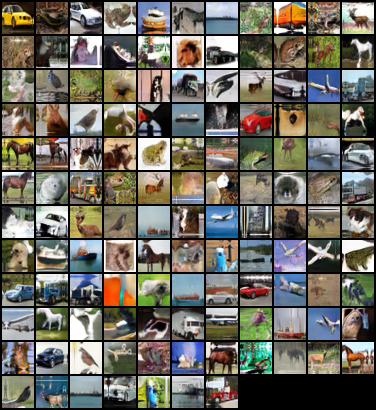}
    \caption{Unconditional CIFAR10 Samples. Left: samples from the generator and right: samples after 100 steps if MCMC refinement using MALA.}
    \label{fig:jem_samp10}
\end{figure}

\begin{figure}[h]  
\centering
\includegraphics[width=.48\textwidth]{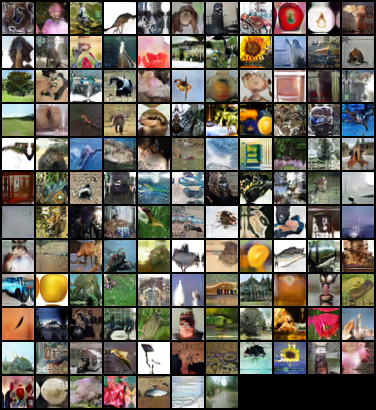}
\hspace{1em}
\includegraphics[width=.48\textwidth]{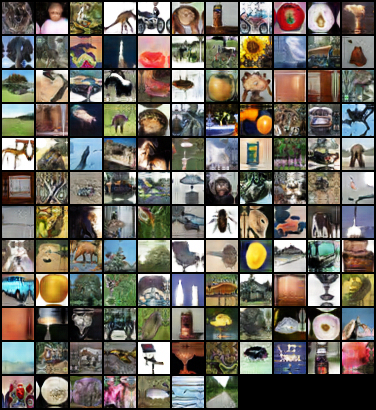}
    \caption{Unconditional CIFAR100 Samples. Left: samples from the generator and right: samples after 100 steps if MCMC refinement using MALA.}
    \label{fig:jem_samp100}
\end{figure}

Our generator $q_\phi(x)$ is trained to approximate our EBM $p_\theta(x)$. After training, the samples from the generator are of high quality (see Figures \ref{fig:jem_samp10} and \ref{fig:jem_samp100}, left) but they are not exactly samples from $p_\theta(x)$. We can use MCMC sampling to improve the quality of these samples. We use a simple MCMC refinement procedure based on the Metropolis Adjusted Langevin Algorithm~\citep{besag1994comments} applied to an expanded state-space defined by our generator and perform the Accept/Reject step in the data space. 

We can reparameterize a generator sample $x\sim q_\phi(x)$ as a function $x(z, \epsilon) = g_\psi(z) + \epsilon \sigma$, and we can define an unnormalized density over  $\{z, \epsilon\}$, $\log h(z, \epsilon) \equiv f_\theta(g_\psi(z) + \epsilon \sigma) - \log Z(\theta, \phi)$ which is the density (under $p_\theta(x)$) of the generator sample.

Starting with an initial sample $z_0, \epsilon_0 \sim \mathcal{N}(0, I)$ we define the proposal distribution
\begin{align}
    p(z_t|z_{t-1}, \epsilon_{t-1}) &=  \mathcal{N}\left(z_{t-1} + \frac{\delta}{2}\nabla_{z_{t-1}} \log h(z_{t-1}, \epsilon_{t-1}), \delta^2 I \right)\nonumber\\
     p(\epsilon_t|z_{t-1}, \epsilon_{t-1}) &=  \mathcal{N}\left(z_{t-1} + \frac{\delta}{2}\nabla_{\epsilon_{t-1}} \log h(z_{t-1}, \epsilon_{t-1}), \delta^2 I \right)\nonumber\\
     p(z_t, \epsilon_t|z_{t-1}, \epsilon_{t-1}) &=  p(z_t|z_{t-1}, \epsilon_{t-1})p(\epsilon_t|z_{t-1}, \epsilon_{t-1})\nonumber
\end{align}
and accept a new sample with probability
\begin{align}
    \min\left[ \frac{h(z_t, \epsilon_t)p(z_t, \epsilon_t|z_{t-1}, \epsilon_{t-1})}{ h(z_{t-1}, \epsilon_{t})p(z_{t-1}, \epsilon_{t-1}|z_{t}, \epsilon_{t}) }, 1\right]\nonumber.
\end{align}

Here, $\delta$ is the step-size and is a parameter of the sampler. We tune $\delta$ with a burn-in period to target an acceptance rate of $0.57$.

\begin{figure}[h]  
\centering
\includegraphics[width=.4\textwidth]{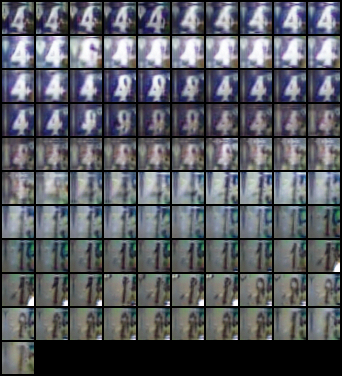}
\hspace{1em}
\includegraphics[width=.4\textwidth]{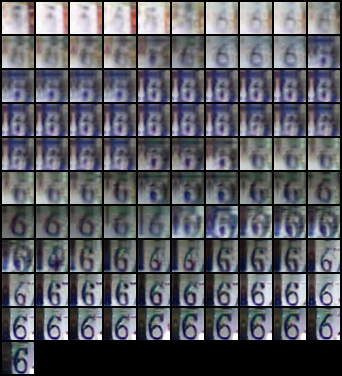}
    \caption{Visualization of our MALA-inspired sample refinement procedure. Samples come from JEM model trained on SVHN. Chains progress to the right and down. Each image is a consecutive step, no sub-sampling is done.}
    \label{fig:mcmc}
\end{figure}

A similar sampling procedure was proposed in \citet{kumar2019maximum} and \citet{che2020your} and in both works was found to improve sample quality. In all experiments, 

We clarify that this procedure is \emph{not} a valid MCMC sampler for $p_\theta(x)$ due to the augmented variables and the change in density of $g_\psi$ which are not corrected for. The density of the samples will be a combination of $p_\theta(x)$ and $q_\phi(x)$. As the focus of this work was training and not sampling/generation, we leave the development of more correct generator MCMC-sampling to future work. Regardless, we find this procedure to improve visual sample quality. In Figure~\ref{fig:mcmc} we visualize a sampling chain using the above method applied to our JEM model trained on SVHN. 

\subsubsection{Image Quality Metrics}
We present results on Inception Score~\citep{salimans2016improved} and Frechet Inception Distance~\citep{heusel2017gans}. These metrics are notoriously fickle and different repositories are known to give very different results~\citep{grathwohl2019your}. For these evaluations we generate 12,800 samples from the model and (unless otherwise stated) refine the samples with 100 steps of our latent-space MALA procedure (Appendix~\ref{app:mala}). The code to generate our reported FID comes from \href{https://github.com/bioinf-jku/TTUR/blob/master/fid.py}{this} publicly available repository. The code to generate our reported Inception Score can be found \href{https://github.com/openai/ebm_code_release/blob/master/inception.py}{here}.

\subsection{Semi-supervised learning on Tabular Data}
\label{app:ssl}

\subsubsection{Data}

We provide details about each of the datasets used for the experiments in Section \ref{sec:ssl}. HEPMASS\footnote{\url{http://archive.ics.uci.edu/ml/datasets/HEPMASS}} is a dataset obtained from a particle accelerator where we must classify signal from background noise. CROP\footnote{\url{https://archive.ics.uci.edu/ml/datasets/Crop+mapping+using+fused+optical-radar+data+set}} is a dataseset for classifying crop types from optical and radar sensor data. HUMAN\footnote{\url{https://archive.ics.uci.edu/ml/datasets/Human+Activity+Recognition+Using+Smartphones}} is a dataset for human activity classification from gyroscope data. MNIST is an image dataset of handwritten images, treated here as tabular data.

\begin{table}[h]
    \centering
    \begin{tabular}{@{}l|llll@{}}
        \toprule
        Dataset             & HEPMASS   & CROP      & HUMAN     & MNIST   \\ 
        \midrule
        Features            & 15              & 174     & 523   & 784 \\
        Examples            & 6,300,000         & 263,926 & 6,617 & 60,000 \\
        Classes             & 2               & 7       & 6     & 10 \\
        Max class \%        & 50.0            & 26.1    & 19.4  & 11.2 \\
        \bottomrule
    \end{tabular}
    \caption{Basic information about each tabular dataset.}
    \label{tab:ssl_data}
\end{table}

For HEPMASS and HUMAN data we remove features which repeat the same exact same value more than 5 times.
For CROP data we remove features which have covariation greater than 1.01.
For MNIST we linearly standardize features to the interval $[-1, 1]$.

We take a random 10\% subset of the data to use as a validation set.

\subsubsection{Training}

We use the same architecture for all experiments and baselines. It has 6 layers of hidden units with dimensions $[1000, 500, 500, 250, 250, 250]$ and a Leaky-ReLU nonlinearity with negative-slope $.2$ between each layer of hidden units. The only layers which change between datasets are the input layer and the output layer which change according to the number of features and number of classses respectively.

The training process for semi-supervised learning is similar to JEM with an additional objective commonly used in semi-supervised learning:
\begin{align}
    \log p_\theta(x, y) = \alpha\log p_\theta(y\mid x) + \log p_\theta(x) + \beta H(p_\theta(y\mid x))
\end{align}

where $H(p(y|x))$ is the entropy of the predictive distribution over the labels.

For all models we report the accuracy the model converged to on the held-out validation set. We report the average taken over three training runs with different seeds.

We use equal learning rates for the energy model, generator, and the entropy estimator. We tune the learning rate and decay schedule for supervised models on the full-set of labels and 10 labels per class.

On \methodname{} we tune the learning rate and decay schedule, the weighting of the entropy regularization $\lambda$ and the weighting of the entropy of classification outputs $\beta$.

For VAT we tune the perturbation size $\epsilon \in \{.01, .1, 1, 3, 10\}$. All other hyperparameters were fixed according to tuning on \methodname{}.

For MEG we used the hyperparameters tuned according to \methodname{}.

For JEM we tune the number of MCMC steps in the range $\kappa \in \{20, 40, 80\}$. We generate samples using SGLD with stepsize 1 and noise standard deviation $0.01$ as in \citet{grathwohl2019your}.

\begin{table}[h]
    \centering
    \begin{tabular}{@{}l|llll@{}}
        \toprule
        Dataset             & HEPMASS   & CROP      & HUMAN     & MNIST   \\ 
        \midrule
        Learning rate       & $10^{-5}$  & $10^{-4}$ & $10^{-5}$   & $10^{-5}$ \\
        Number of epochs    & $1$     & $20$    & $800$   & $200$ \\
        Batch size          & $64$    & $64$    & $64$    & $64$ \\
        Decay rate          & $.3$    & $.3$    & $.3$    & $.3$ \\
        Decay epochs        & $\{.05, .1\}$ & $\{15, 17\}$ & $\{200, 300, 400, 600\}$  & $\{150, 175\}$ \\
        $z$ (latent) dimension & $10$ & $128$ & $128$ & $128$ \\
        $\lambda$       & $10^{-4}$ & $10^{-4}$ & $10^{-4}$ & $10^{-4}$  \\
        $\alpha$             & $1$ & $1$ & $10$ & $1$ \\
        $\beta$   & $1$ & $1$ & $1$ & $1$\\
        $\epsilon$ (VAT)  & $.1$ & $3$ & $.1$ & $3$ \\
        $\kappa$ (JEM)            & $20$ & $20$ & $40$ & $80$ \\
        \bottomrule
    \end{tabular}
    \caption{Hyperparameters for semi-supervised learning on tabular data}
    \label{tab:ssl_hp}
\end{table}

\newpage
\section{Additional Results}

\subsection{Training Mixtures of Gaussians}
\label{app:mog}
We present additional results training mixure of Gaussian models using \methodname{} and PCD. Each model consits of 100 diagonal Gaussians and mixing weights. Our experimental setup and hyper-parameter search was identical to that presented in Appendiex \ref{app:toy}. We see in Table~\ref{tab:ml_toy} that \methodname{} outperforms PCD. 
\begin{table}[h]
    \centering
    \begin{tabular}{@{}l|l|lllll@{}}
    \toprule
     & \multirow{2}{*}{Max. Likelihood} & \multicolumn{4}{c}{\methodname{}} &  \multirow{2}{*}{PCD} \\
     &                 &    $\lambda = 0.0$    & $\lambda = .01$& $\lambda = 0.1$ & $\lambda = 1.0$                    & \\
    \midrule
    Moons & -2.32 & -3.87  & -3.63 & -3.10 & \textbf{-2.58} & -3.82\\
    Circles & -3.17 & -3.58  & -3.64 & -3.74 & \textbf{-3.40} & -4.24\\
    Rings & -2.83 &  -3.5  & -3.44 & -3.24 & \textbf{-3.17} & -4.13\\
    \bottomrule
    \end{tabular}
    \caption{Fitting a mixture of 100 diagonal Gaussians using ML, MCMC approximate ML and generator approximate ML.}
    \label{tab:ml_toy}
\end{table}

\subsection{Samples from SSL Models}
\begin{figure}[h]  
\centering
\includegraphics[width=.2\textwidth]{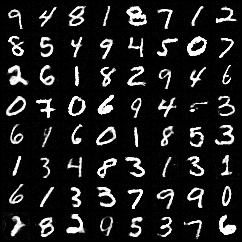}
\hspace{1em}
\includegraphics[width=.2\textwidth]{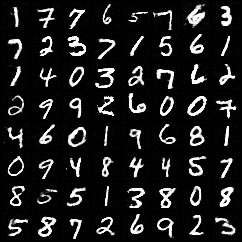}
    \caption{Unconditional MNIST Samples. Left: samples from the generator and right: samples after 100 steps if MCMC refinement using MALA.}
    \label{fig:jem_ssl}
\end{figure}

We present some samples from our semi-supervised MNIST models in Figure~\ref{fig:jem_ssl}.

\subsection{Hybrid Modeling}
We present an extended Table~\ref{tab:jem_fid} with Inception Score~\cite{salimans2016improved} and which includes more comparisons.

\begin{table}[h]
\centering
\begin{tabular}{@{}l|ll@{}}
    \toprule
    Model & FID & IS\\
    \midrule
     JEM & 38.4 & 8.76\\
     HDGE & 37.6 & 9.19\\
     SNGAN~\citep{miyato2018spectral} & 25.50 & 8.59\\
     NCSN~\citep{song2019generative} & 23.52 & 8.91\\
     ADE~\citep{dai2019exponential} & N/A & 7.55 \\
     IGEBM~\citep{du2019implicit} &  37.9 & 8.30 \\
     Glow~\citep{kingma2018glow} & 48.9 & 3.92\\
     FCE~\citep{gao2020flow} & 37.3 & N/A\\
     \midrule
     \methodname{} $\alpha = 100$  & 30.5 & 8.11\\
     \methodname{} $\alpha = 1$  & 27.5 & 8.00\\
     \methodname{} $\alpha = 1$ (generator samples) & 32.4 & 7.34 \\
    \bottomrule
    \end{tabular}
\end{table}

\newpage

\subsection{Unconditional Image Generation: CIFAR10}
\label{app:samp}

\subsection{Unconditional Image Generation: CIFAR100}
\label{app:samp}
\newpage

\subsection{Conditional Image Generation: CIFAR10 and CIFAR100}
\begin{figure}[h]  
\centering
\includegraphics[width=.8 \textwidth]{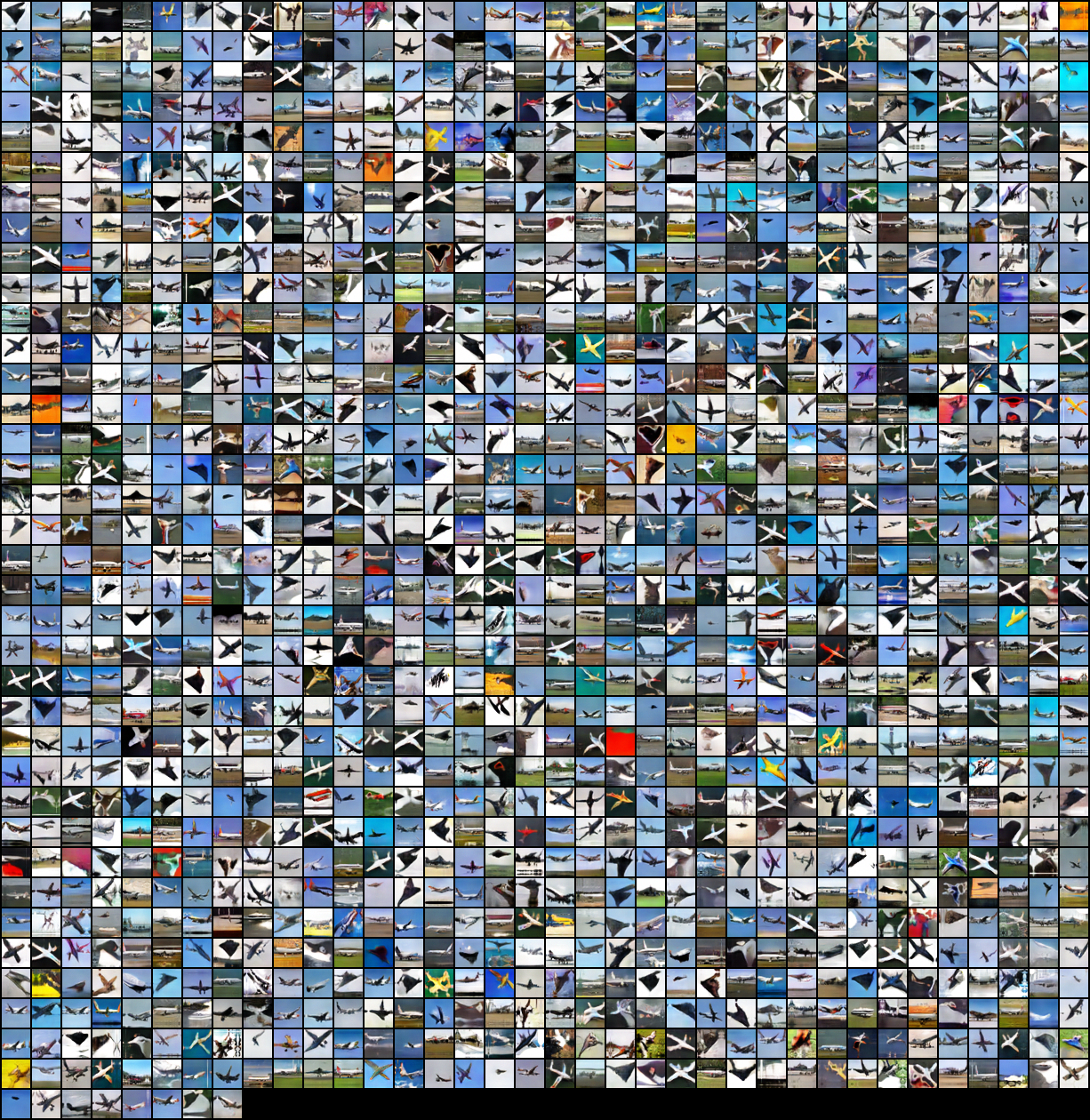}
    \caption{Class-conditional samples from CIFAR10}
    \label{fig:jem_samp}
\end{figure}

\begin{figure}[h]  
\centering
\includegraphics[width=.8 \textwidth]{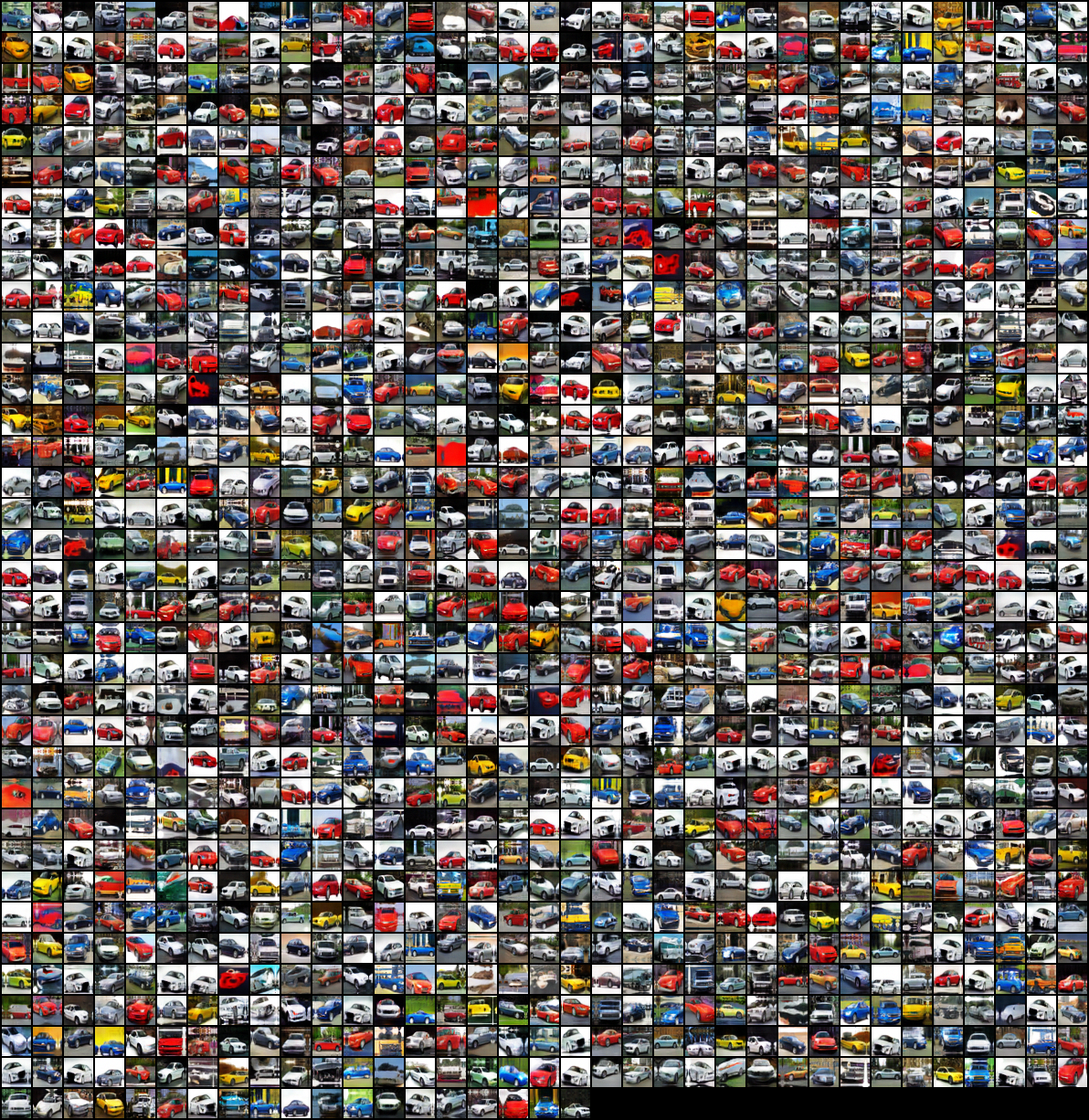}
    \caption{Class-conditional samples from CIFAR10}
    \label{fig:jem_samp}
\end{figure}

\begin{figure}[h]  
\centering
\includegraphics[width=.8 \textwidth]{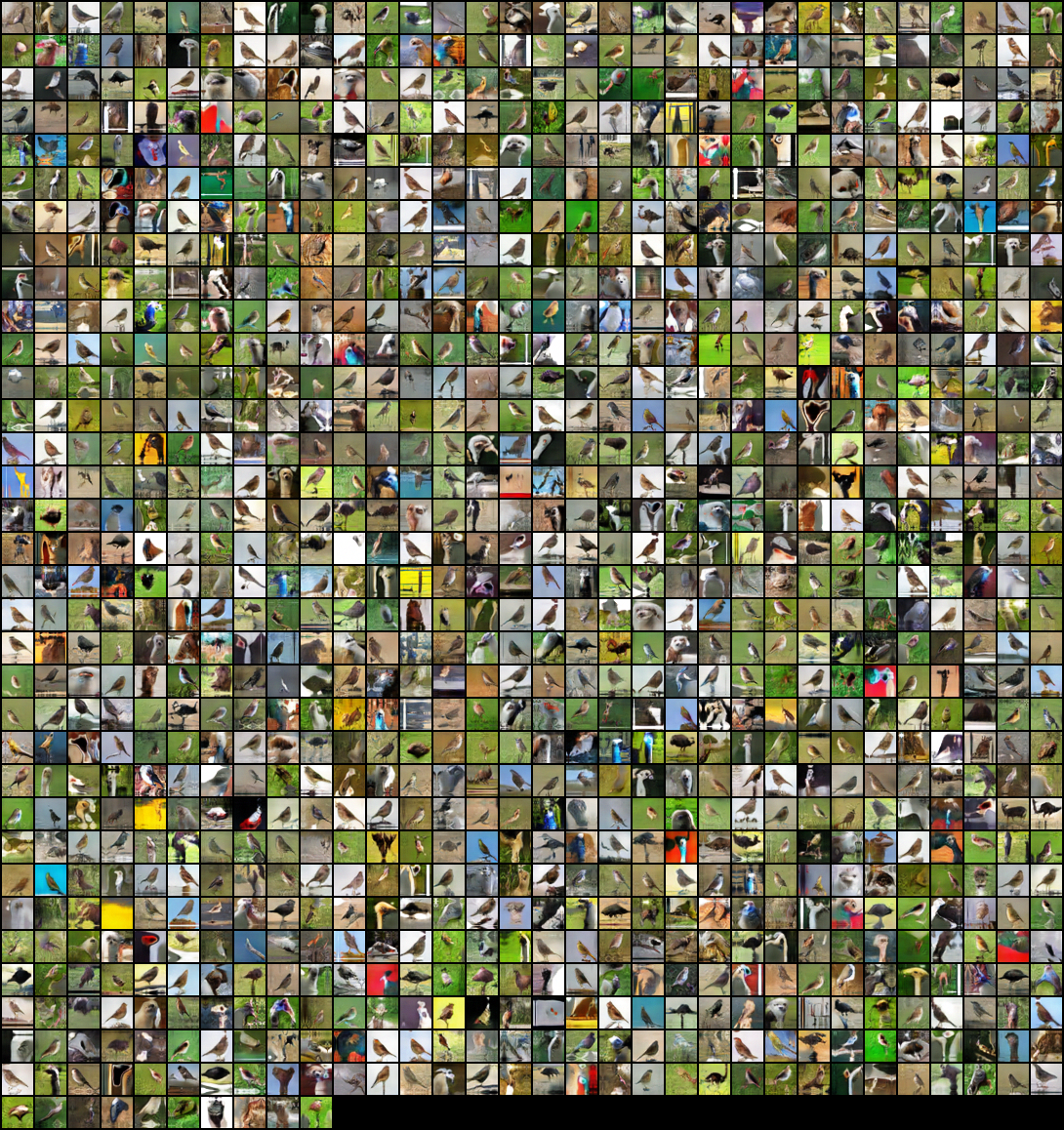}
    \caption{Class-conditional samples from CIFAR10}
    \label{fig:jem_samp}
\end{figure}

\begin{figure}[h]  
\centering
\includegraphics[width=.8 \textwidth]{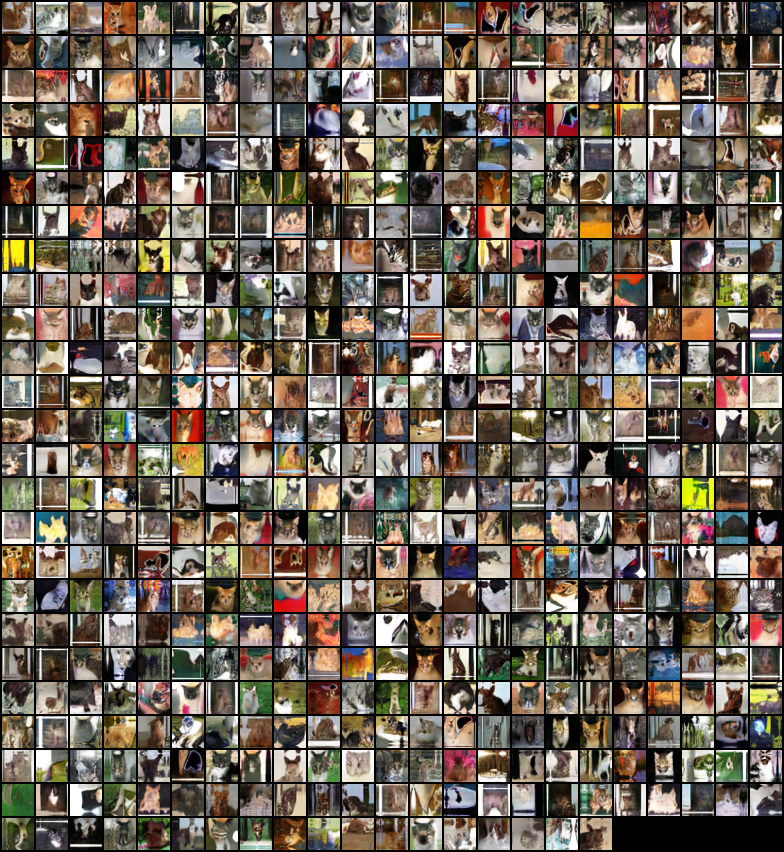}
    \caption{Class-conditional samples from CIFAR10}
    \label{fig:jem_samp}
\end{figure}

\begin{figure}[h]  
\centering
\includegraphics[width=.8 \textwidth]{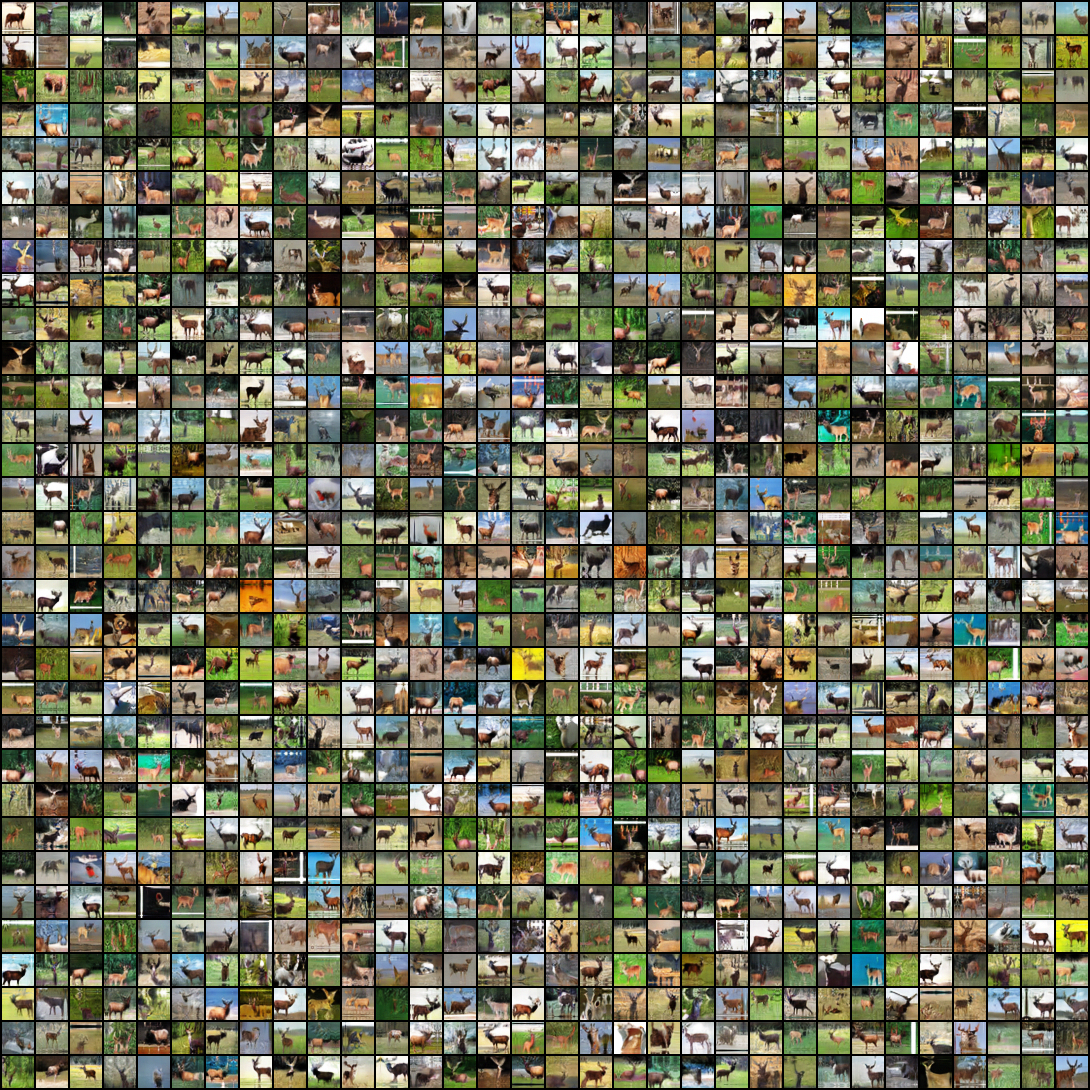}
    \caption{Class-conditional samples from CIFAR10}
    \label{fig:jem_samp}
\end{figure}

\begin{figure}[h]  
\centering
\includegraphics[width=.8 \textwidth]{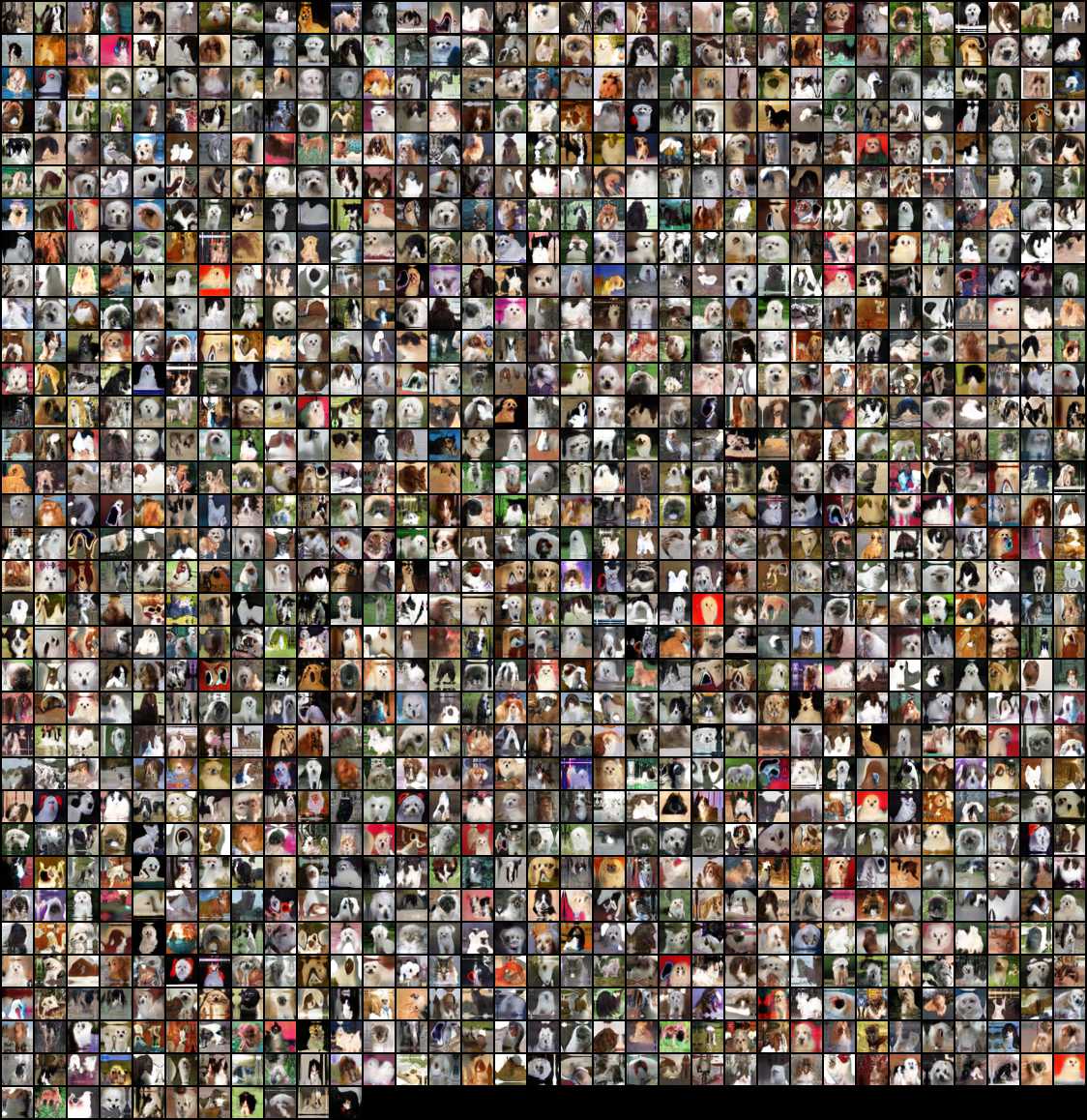}
    \caption{Class-conditional samples from CIFAR10}
    \label{fig:jem_samp}
\end{figure}

\begin{figure}[h]  
\centering
\includegraphics[width=.8 \textwidth]{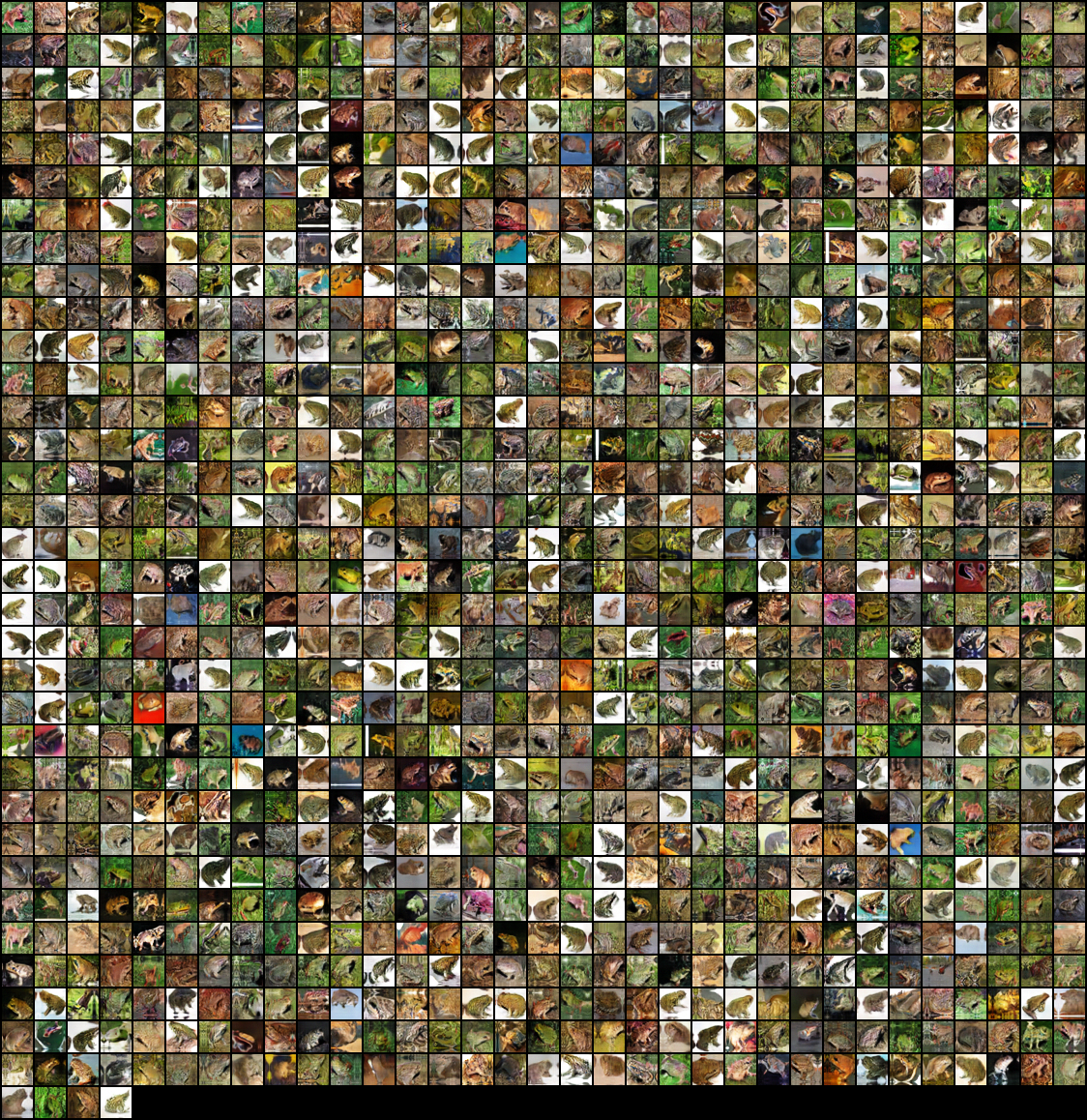}
    \caption{Class-conditional samples from CIFAR10}
    \label{fig:jem_samp}
\end{figure}

\begin{figure}[h]  
\centering
\includegraphics[width=.8 \textwidth]{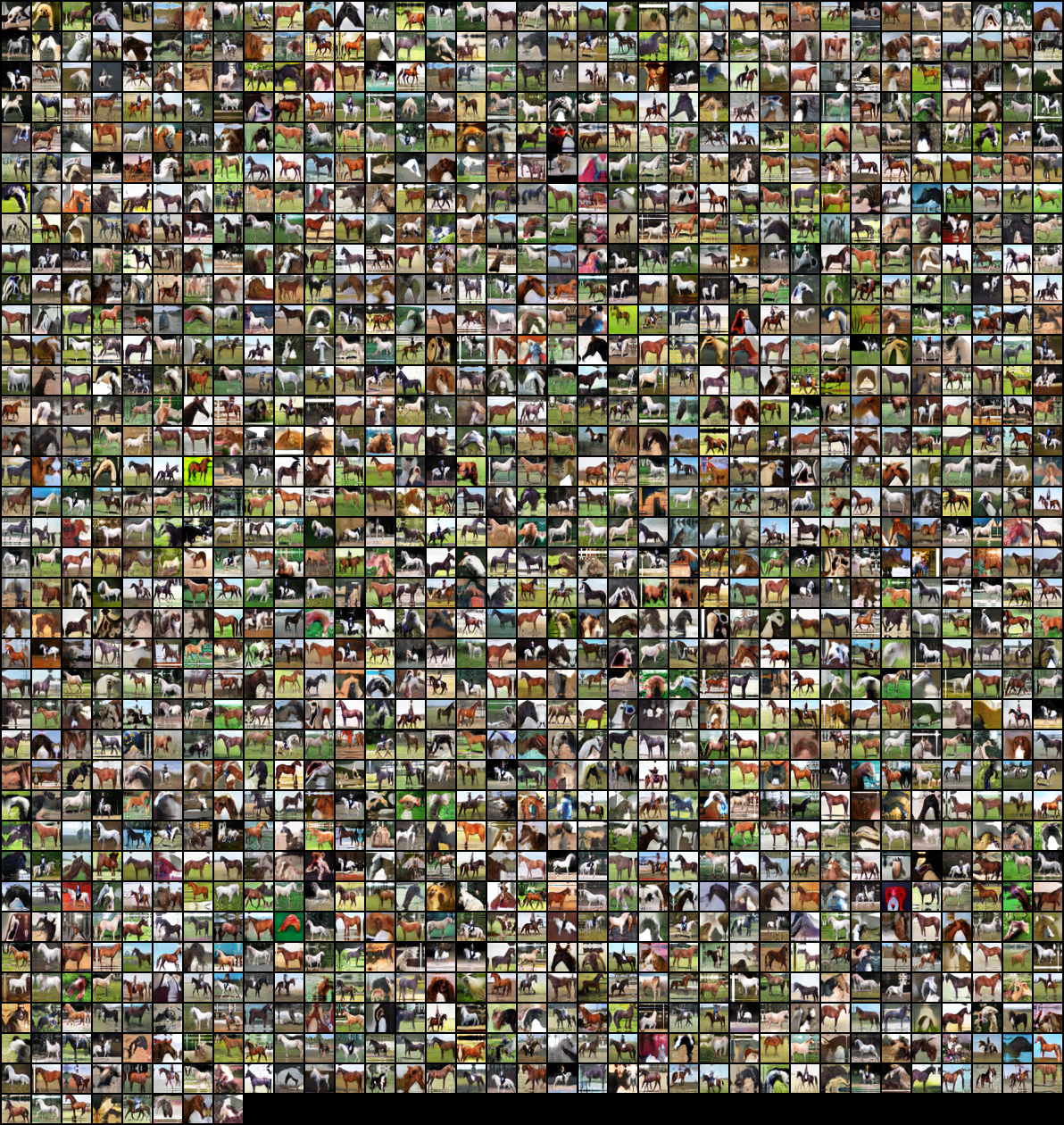}
    \caption{Class-conditional samples from CIFAR10}
    \label{fig:jem_samp}
\end{figure}

\begin{figure}[h]  
\centering
\includegraphics[width=.8 \textwidth]{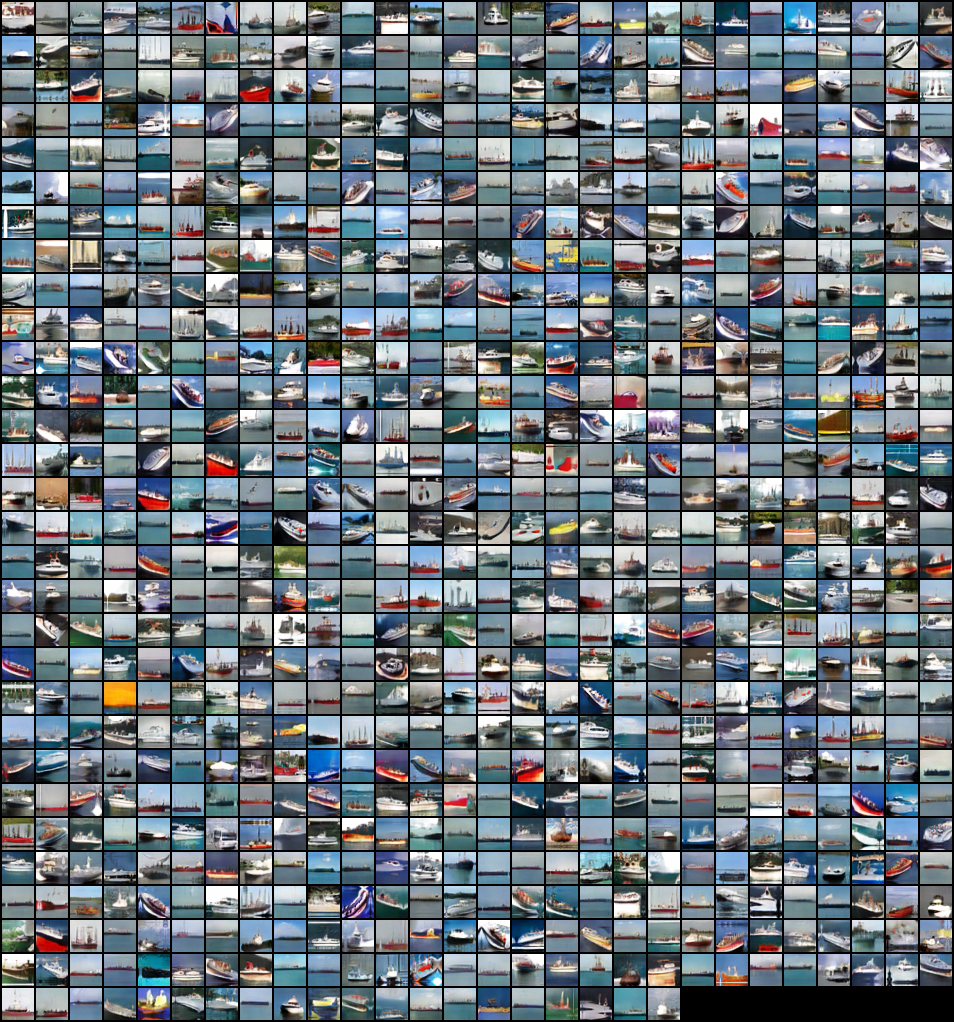}
    \caption{Class-conditional samples from CIFAR10}
    \label{fig:jem_samp}
\end{figure}

\begin{figure}[h]  
\centering
\includegraphics[width=.8 \textwidth]{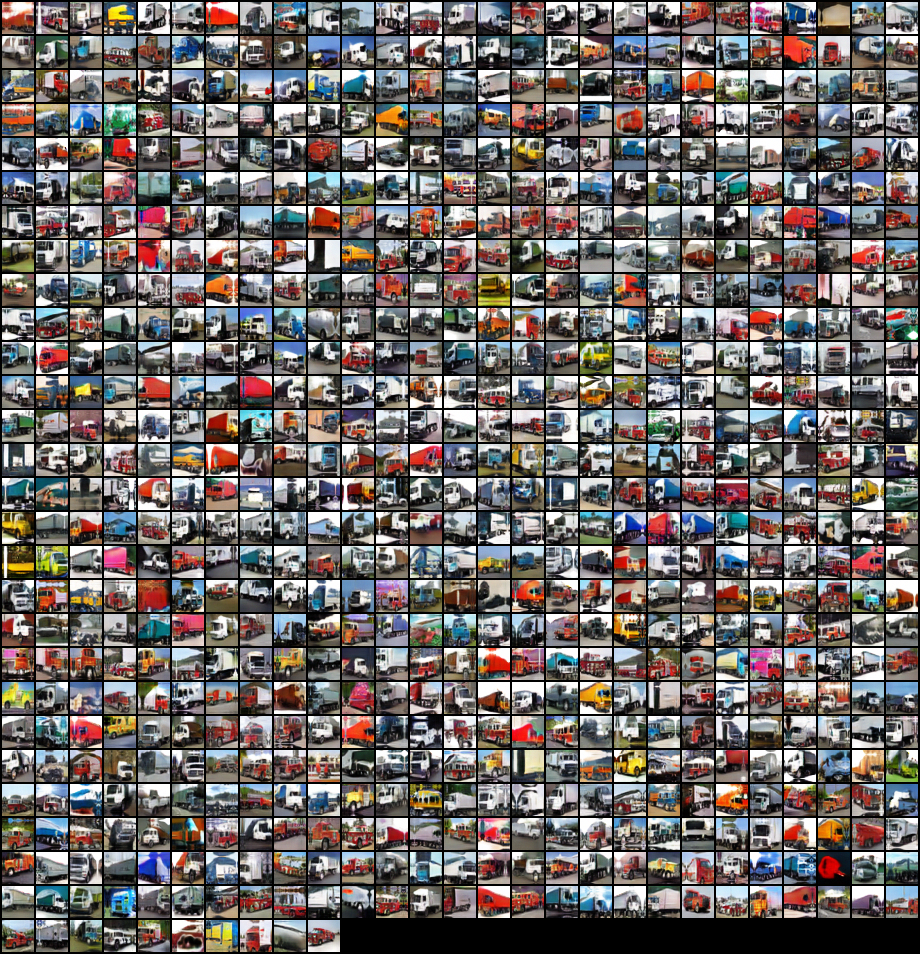}
    \caption{Class-conditional samples from CIFAR10}
    \label{fig:jem_samp}
\end{figure}

\newpage\newpage

\begin{figure}[h!]  
\centering
\includegraphics[width=.4 \textwidth]{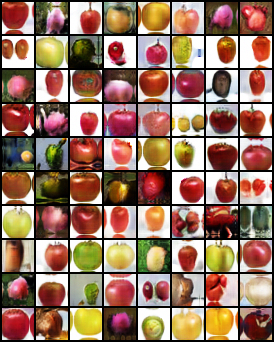}
\hspace{1em}
\includegraphics[width=.4 \textwidth]{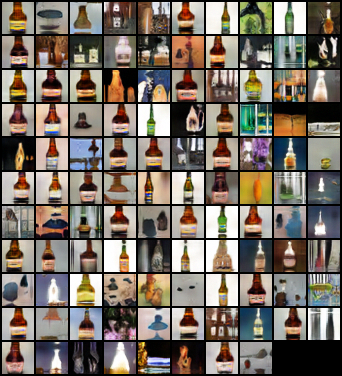}
    \caption{Class-conditional samples from CIFAR100}
    \label{fig:jem_samp}
\end{figure}

\begin{figure}[h!]  
\centering
\includegraphics[width=.4 \textwidth]{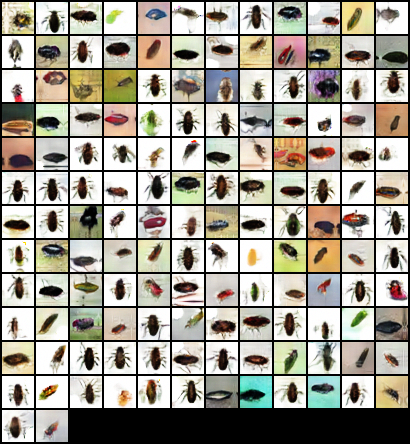}
\hspace{1em}
\includegraphics[width=.4 \textwidth]{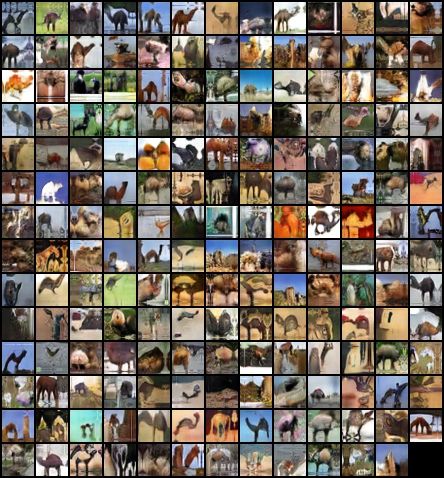}
    \caption{Class-conditional samples from CIFAR100}
    \label{fig:jem_samp}
\end{figure}

\begin{figure}[h!]  
\centering
\includegraphics[width=.4 \textwidth]{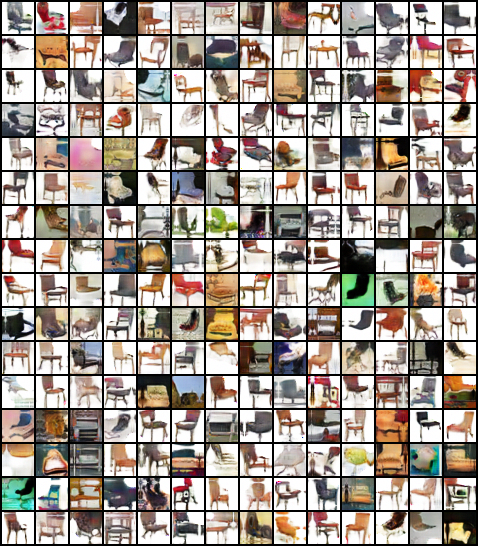}
\hspace{1em}
\includegraphics[width=.4 \textwidth]{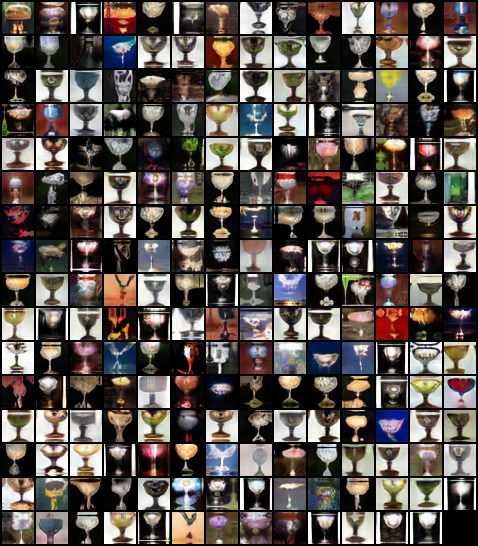}
    \caption{Class-conditional samples from CIFAR100}
    \label{fig:jem_samp}
\end{figure}

\begin{figure}[h!]  
\centering
\includegraphics[width=.4 \textwidth]{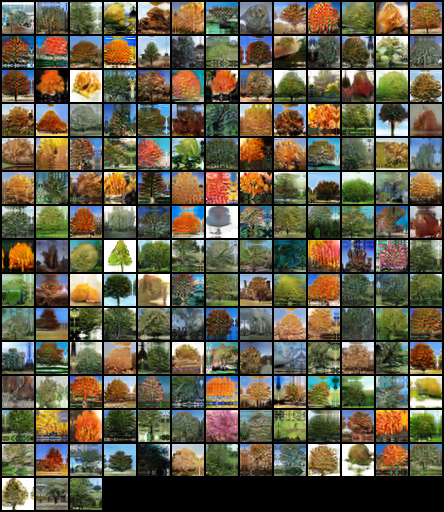}
\hspace{1em}
\includegraphics[width=.4 \textwidth]{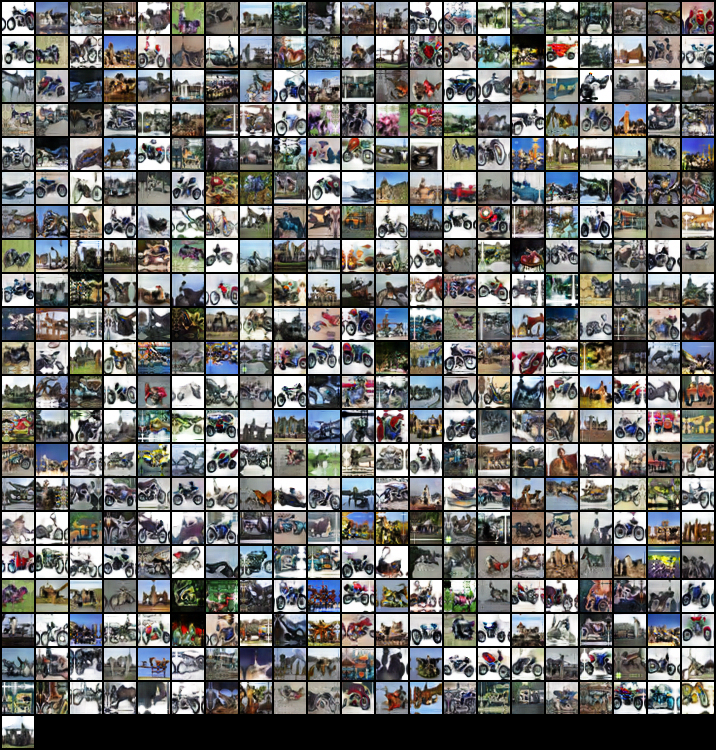}
    \caption{Class-conditional samples from CIFAR100}
    \label{fig:jem_samp}
\end{figure}

\begin{figure}[h!]  
\centering
\includegraphics[width=.4 \textwidth]{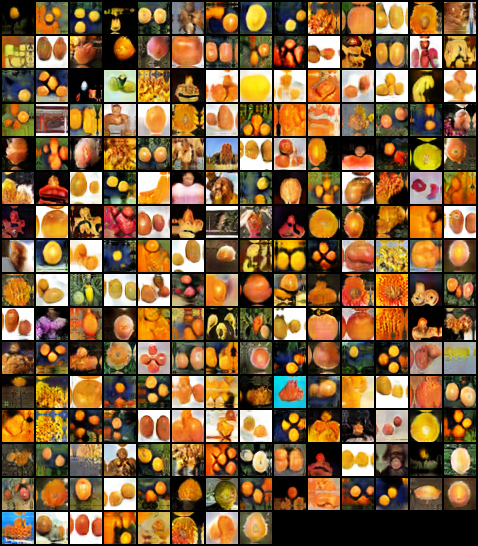}
\hspace{1em}
\includegraphics[width=.4 \textwidth]{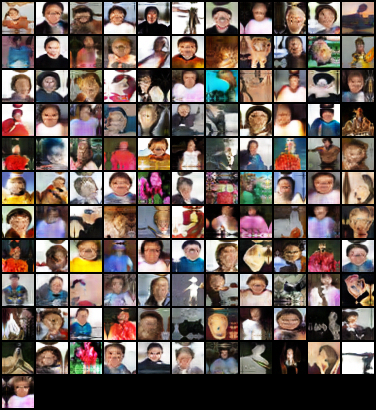}
    \caption{Class-conditional samples from CIFAR100}
    \label{fig:jem_samp}
\end{figure}

\begin{figure}[h!]  
\centering
\includegraphics[width=.4 \textwidth]{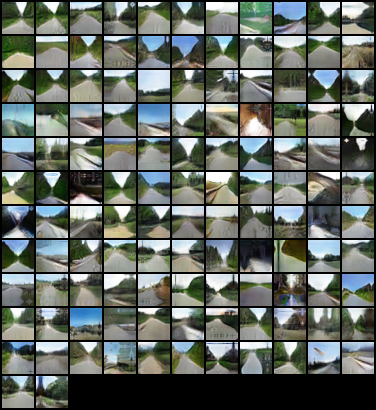}
\hspace{1em}
\includegraphics[width=.4 \textwidth]{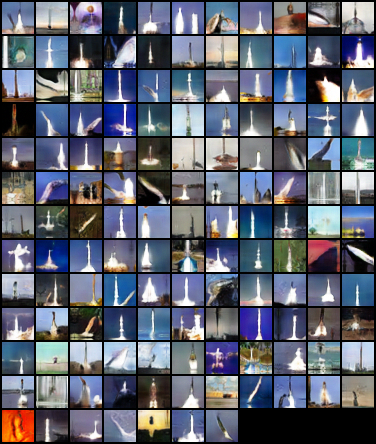}
    \caption{Class-conditional samples from CIFAR100}
    \label{fig:jem_samp}
\end{figure}

\begin{figure}[h!]  
\centering
\includegraphics[width=.4 \textwidth]{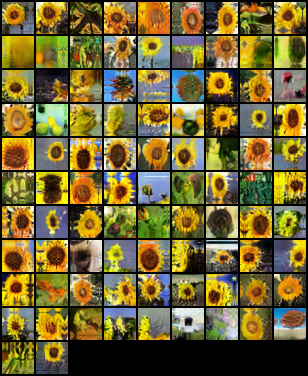}
\hspace{1em}
\includegraphics[width=.4 \textwidth]{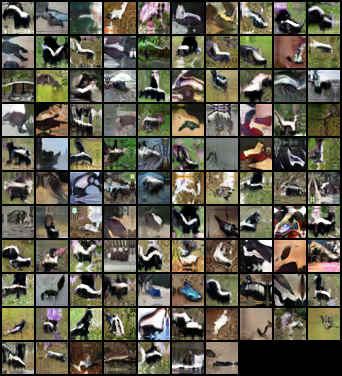}
    \caption{Class-conditional samples from CIFAR100}
    \label{fig:jem_samp}
\end{figure}

\end{document}